\documentclass[10pt]{article}

\usepackage[a4paper,margin=2.2cm]{geometry}
\usepackage[T1]{fontenc}
\usepackage[utf8]{inputenc}
\usepackage{lmodern}
\usepackage{microtype}
\usepackage{graphicx}
\usepackage{booktabs}
\usepackage{array}
\usepackage{makecell}
\usepackage{amsmath,amssymb}
\usepackage{siunitx}
\usepackage[font=small,labelfont=bf]{caption}
\usepackage[numbers,sort&compress]{natbib}
\usepackage{xcolor}
\usepackage[colorlinks=true,allcolors=blue]{hyperref}
\renewcommand{\topfraction}{0.9}
\renewcommand{\bottomfraction}{0.8}
\renewcommand{\textfraction}{0.07}
\renewcommand{\floatpagefraction}{0.8}

\graphicspath{{figures/}}

\DeclareSIUnit\milliamperepercmsquared{mA\,cm^{-2}}

\title{Physics-residual machine learning predicts oxygen-evolution catalyst activity beyond the training range from sparse polarization measurements}
\author{\begin{minipage}{0.92\textwidth}\centering
Yong-Woon~Kim$^{1,2}$, Jihyeok~Lee$^{3}$, Sungtae~Park$^{4}$, Sooseok~Choi$^{3}$, Yung-Cheol~Byun$^{5,*}$\\[0.6em]
{\small $^{1}$Department of Computer Engineering, Jeju National University, Jeju 63243, South Korea\\
$^{2}$Green Hydrogen Global Leading Research Center, Jeju National University, Jeju 63243, South Korea\\
$^{3}$Faculty of Applied Energy System, Jeju National University, Jeju 63243, South Korea\\
$^{4}$Nuclear-Hydrogen Convergence Center, Research and Development Section, Korea Hydro and Nuclear Power Co. Ltd., Daejeon 34101, South Korea\\
$^{5}$Department of Computer Engineering, Major of Electronic Engineering, Food Tech Center (FTC), Jeju National University, Jeju 63243, South Korea\\[0.35em]
Email addresses: ywkim@jejunu.ac.kr (Y.-W. Kim), sooseok@jejunu.ac.kr (S. Choi), ycb@jejunu.ac.kr (Y.-C. Byun)\\
$^{*}$Corresponding author: Yung-Cheol Byun}
\end{minipage}}
\date{}

\begin{document}
\maketitle

\begin{abstract}
\noindent Discovery campaigns for oxygen evolution reaction catalysts repeatedly choose, make and measure
catalysts. High-throughput platforms stop polarization curves below potentials that damage the
catalyst, so the endpoint, the activity at a target potential or current density, often lies beyond
the measured window, and the catalysts of most interest are more active than any measured before.
Existing methods do not predict these endpoints accurately when few or no endpoints of a new library
have been measured. Here we present physics-residual machine learning (PR-ML), which predicts each
endpoint as the sum of a Tafel term, computed from the catalyst's own measured curve with an estimated
slope, and a residual term learned from labelled catalysts. In twelve Ni--Pd--Pt--Ru thin-film
libraries, the current density at 1.70~V$_\mathrm{RHE}$ was predicted from the currents at 1.40 and
1.55~V$_\mathrm{RHE}$. Fitted only on earlier libraries, with ridge regression as the residual
learner, PR-ML predicted the Ni--Ru library, whose currents mostly exceed theirs, with a mean absolute
error of 0.194~\si{\milliamperepercmsquared}, against 1.330--1.882 for data-driven models. With five
endpoints from the new library and extremely randomized trees as the residual learner, PR-ML gave a
similar error, which the same learner used alone reached only with 20, and identified 63--83\% of the
catalysts more active than the best labelled catalyst, against 2\%. In two independent datasets, this
fraction rose from at most 1\% to 33--95\%. Our approach supplies catalyst selection with accurate
endpoints beyond the measured part of each curve and above all earlier measurements.
\end{abstract}

\section*{Introduction}\phantomsection\label{sec:introduction}

Water electrolysis stores electrical energy as hydrogen, and the oxygen evolution reaction
(OER) at the anode accounts for a large part of the energy lost in that conversion
\cite{Rossmeisl2007,Man2011}. New OER catalysts are found by repeating three steps: choosing
which catalysts to make, making them, and measuring them. Combinatorial thin-film libraries
and other high-throughput experiments prepare hundreds to thousands of compositions together
and measure them under identical conditions \cite{Thelen2025OER,Xu2024}, and
machine-learning-guided synthesis and robotic platforms repeat the making and measuring of
new catalysts \cite{DASH2025,FastCat2025,Zhang2025CRESt}. Sequential learning, active
learning and Bayesian optimization then use the measured activities to choose the catalysts
to make next \cite{Rohr2020Sequential,Moon2024ActiveOER,Thelen2025OER}. Every later decision
of a discovery campaign is therefore based on the activities obtained in its measurement
step.

OER catalysts are compared by the current density they reach at a given potential or by the
overpotential they need to reach a given current density, with
10~\si{\milliamperepercmsquared} as the common reference current density \cite{McCrory2013}.
We call the current density at a target potential, or the overpotential at a target current
density, the endpoint. High-throughput platforms record the polarization curve of each
catalyst within a fixed potential window; for the twelve Ni--Pd--Pt--Ru thin-film libraries
of Thelen et al., the window ends at 1.8~V versus the reversible hydrogen electrode
(V$_\mathrm{RHE}$) \cite{Thelen2025OER}. Higher potentials are avoided because bubbles block
active sites and the catalyst dissolves faster \cite{Zlatar2023}: repeated sweeps to
2.0~V$_\mathrm{RHE}$ caused an irreversible activity loss attributed to dissolution or
passivation of active sites \cite{Petzoldt2021}, and iridium dissolution increases with the
OER rate \cite{Loncar2022}. As a result, the endpoint often lies outside the measured window:
in the twelve Ni--Pd--Pt--Ru libraries, none of the 4,026 measurement areas reached
10~\si{\milliamperepercmsquared} up to 1.8~V$_\mathrm{RHE}$ \cite{Thelen2025OERData}.
Campaigns also aim to find catalysts more active than any measured before
\cite{Rohr2020Sequential}. When each Ni--Pd--Pt--Ru library was predicted from the others,
the largest error occurred for the Ni--Ru library \cite{Thelen2025OER}, whose currents mostly
exceed those of the libraries prepared before it \cite{Thelen2025OERData}. The endpoints used
to select catalysts for further testing therefore often lie beyond the measured part of each
curve, and the catalysts of most interest are more active than any catalyst measured before.

Existing methods address parts of this problem. \textbf{A)} The Tafel equation extends a
measured polarization curve to higher potentials \cite{Stern1957}. It gives the correct
endpoint only when the measured part lies in a region of constant Tafel slope, but the slope
changes with overpotential \cite{Shinagawa2015} and slopes taken from potentiodynamic curves
differ from steady-state slopes \cite{Anantharaj2021TafelPitfalls}, so a standard slope such
as 40, 60 or 120~mV~dec$^{-1}$ applied outside that region gives incorrect endpoints.
\textbf{B)} Data-driven models learn the endpoint from the composition and other descriptors
of catalysts whose endpoints were measured (labelled catalysts). The Ni--Pd--Pt--Ru study used
a Gaussian process (GP) on composition \cite{Thelen2025OER}, and descriptor and graph models
are widely used to screen catalysts \cite{Xu2024,Liu2026CatalystGNN}. Their accuracy
decreases outside the training distribution \cite{Omee2024OOD,MatUQ2026}, especially above
the range of the training labels \cite{KnownUnknowns2025}, and tree ensembles predict
averages of training labels, which cannot exceed the largest label \cite{ExtraTrees2006}.
\textbf{C)} Methods that reduce the number of measurements measure part of a library and
predict the remainder from composition \cite{Thelen2023ActiveLearning,Thelen2025OER}, and
multi-fidelity learning combines many inexpensive measurements with a few expensive ones
through relations learned from data \cite{Fare2022,SabanzaGil2025}. Neither uses a physical
relation between the measured part of a catalyst's curve and its endpoint, so their
predictions depend on endpoint measurements that cover the activity range of the catalysts
being predicted. \textbf{D)} Hybrid models in chemistry and catalysis add a learned
correction to a physically computed value \cite{Ramakrishnan2015DeltaML,Mamun2020,Kim2026PEM}, and
Gaussian processes can use a physical model as their mean function
\cite{Noack2021GP,Ziatdinov2022sGP}; in these models the physical term is computed from
composition, lower-level calculations or the input variables of the search. In
electrochemistry, Thelen et al.\ interpolate Butler--Volmer coefficients, fitted to fully
measured curves, across compositions \cite{Thelen2026SECCM}; this does not extend a measured
curve beyond its window. In electrocatalyst activity screening, these methods therefore do not provide accurate endpoints for the catalysts a campaign most needs to rank: catalysts whose endpoints lie beyond the measured window and above all earlier measurements, in a new library where few or no endpoints have been measured.

Here we predict the endpoint of each catalyst from its own measurements with a
physics-residual machine-learning (PR-ML) model. The prediction is the sum of a Tafel term and a residual term. The
Tafel term, the physical term of the model, extends the value measured at the catalyst's highest potential or current density
to the endpoint with a slope estimated from the measured polarization curves rather than a
standard slope, and the residual term corrects the error that remains where the slope changes
between the measured window and the endpoint (A). Because the prediction is computed from the
catalyst's own measured value, it is not limited to the range of activities in the training
data (B). The residual term is the output of a machine-learning model, the residual learner, fitted
to the difference between the Tafel term and the measured endpoints of labelled catalysts;
because the residual learner fits only this difference, few endpoint measurements are enough, and they
need not cover the activity range of the catalysts being predicted (C). Unlike the physical
term of earlier hybrid models, the Tafel term is computed from the measurements of the
catalyst being predicted (D). PR-ML is designed for the measurement step of each round:
it is fitted either on earlier libraries or on  measured points in the new library,
and provides its predicted endpoints as input to the methods that choose the next catalysts.

In the Ni--Ru library, with no
endpoint measured in that library, and with ridge regression (Ridge) as the residual learner, PR-ML
(Ridge) predicted the current density at 1.70~V$_\mathrm{RHE}$ from currents measured at 1.40 and
1.55~V$_\mathrm{RHE}$ with a mean absolute error (MAE) of 0.194~\si{\milliamperepercmsquared}, about
seven to eight times lower than Ridge used alone (1.330~\si{\milliamperepercmsquared}) and
GP (composition), of the type used in the original study (1.651~\si{\milliamperepercmsquared}). With
the endpoints of five catalysts measured in the new library and extremely randomized trees
(ExtraTrees) as the residual learner, PR-ML (ExtraTrees) had an MAE of
0.199~\si{\milliamperepercmsquared}; ExtraTrees used alone reached a similar error only with 20
labelled catalysts. PR-ML (ExtraTrees) also identified on average 63--83\% of the catalysts more active
than the best labelled catalyst in twelve libraries, against 2\% for ExtraTrees, and in two OER datasets collected independently
in other laboratories, with endpoint labels taken only from the less active 50\% or 70\% of
the catalysts, this fraction rose from at most 1\% to 33--95\% \cite{FastCat2025,DASH2025}.

This work makes three contributions. \textbf{First}, it predicts endpoints beyond the
measured part of each polarization curve and above the activities of all training catalysts,
the prediction a campaign needs when the measured potential window ends before the endpoint.
Whether a new library lies above the training range is not known before its endpoints are
measured; over the Ni--Ru library and six later libraries inside the training range, the
larger of the two errors of PR-ML (Ridge), 0.194~\si{\milliamperepercmsquared}, is lower than that of
every compared model, and for the data-driven models this value is
1.330--1.882~\si{\milliamperepercmsquared}. \textbf{Second}, it shows that the proposed approach keeps its accuracy with different residual learners. With a GP instead of ExtraTrees as the residual learner, the error is similar
(0.208 against 0.199~\si{\milliamperepercmsquared}), and in the two independent datasets, ExtraTrees
or a GP as the residual learner gave a lower error for catalysts above the label range than
the same learner used alone in 68 of 80 comparisons with the same features and labels, and a
higher error in none. \textbf{Third}, it gives rules for campaigns. About five endpoint
measurements in a new library give the accuracy obtained by fitting on earlier libraries, and
the catalysts whose current at 1.55~V$_\mathrm{RHE}$ exceeds the highest
1.55~V$_\mathrm{RHE}$ current of the training libraries can be selected before any endpoint
is measured; for these catalysts the error of PR-ML (Ridge) was 22--73\% lower than that of the Tafel
term, with two to five measured currents per catalyst.

Because every discovery campaign measures the catalysts it makes, our approach can be applied in the measurement step of each round, giving the methods that choose the next catalysts accurate endpoints beyond the measured part of each curve and above all earlier measurements, without measurements at the potentials that damage the catalyst.

\section*{Results}\phantomsection\label{sec:results}

\subsection*{The endpoints of high-throughput OER measurements often lie beyond the measured part of the curve}

We used three public OER datasets. The Ni--Pd--Pt--Ru dataset contains twelve thin-film
composition libraries, numbered in the order in which they were prepared, with 322--340
measurement areas per library and 4,026 in total; each area was measured by linear sweep
voltammetry from 1.0 to 1.8~V$_\mathrm{RHE}$ \cite{Thelen2025OER,Thelen2025OERData}, and each
measurement area is treated as one catalyst. From FastCat \cite{FastCat2025,FastCatData2025} and
DASH \cite{DASH2025,DASHData2025} we used the overpotentials at 1, 2, 5, 10, 20 and
50~\si{\milliamperepercmsquared} of the 89 FastCat catalysts and of the 135 of 251 DASH catalysts
whose curves reach all six current densities (\hyperref[sec:methods]{Methods}).

\begin{figure}[!tbp]
\centering
\includegraphics[width=\textwidth]{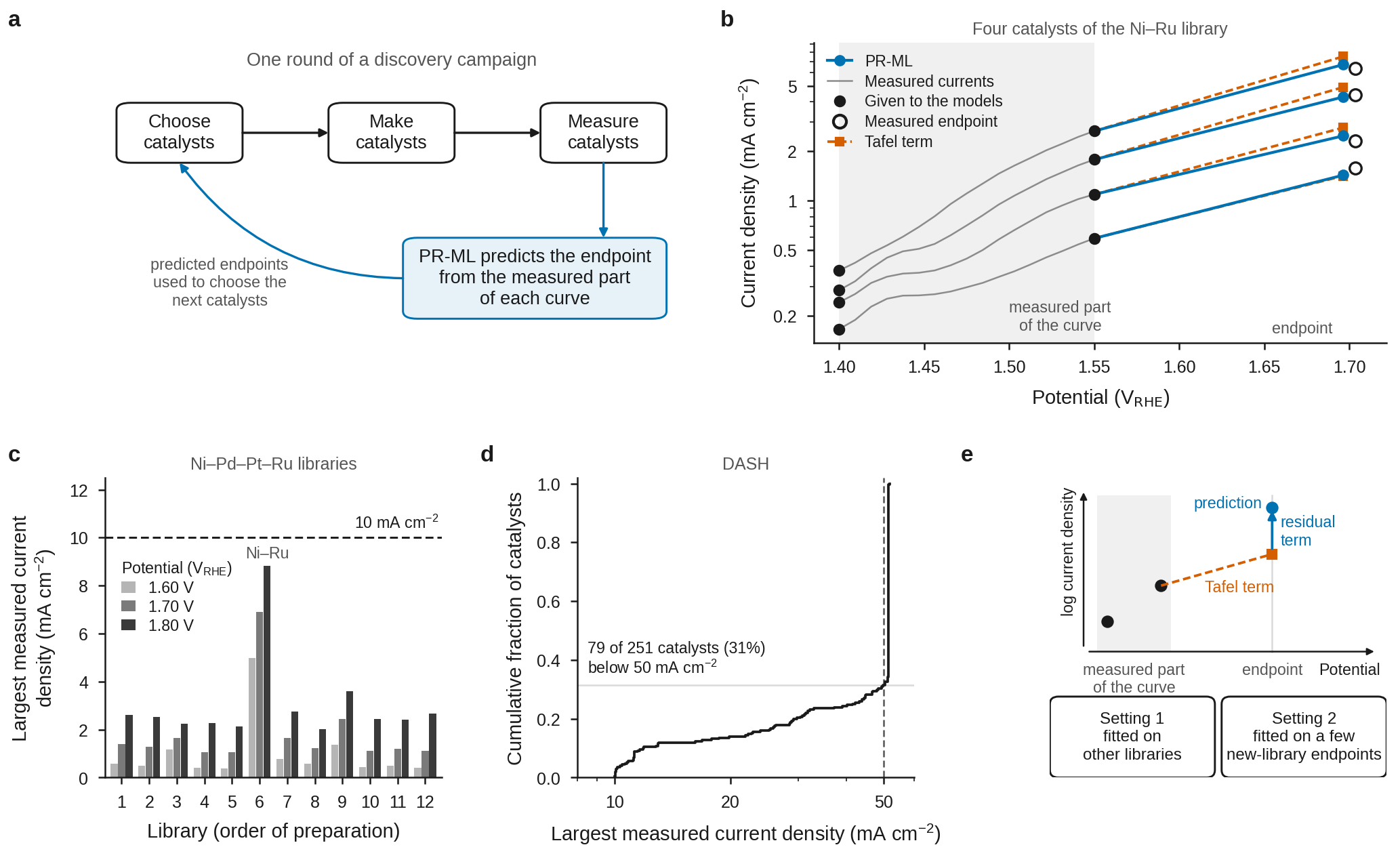}
\caption{\textbf{Endpoints in high-throughput OER measurements and physics-residual machine learning (PR-ML).}
\textbf{a}, One round of a discovery campaign: catalysts are chosen, made and measured, and PR-ML, applied in the measurement step, provides predicted endpoints that are used
to choose the next catalysts. \textbf{b}, Four catalysts of the Ni--Ru library. Grey lines, currents
measured at 17 potentials between 1.40 and 1.55~V$_\mathrm{RHE}$ (grey band); filled black circles,
the currents at 1.40 and 1.55~V$_\mathrm{RHE}$ given to the models; dashed lines and squares, Tafel
term; solid lines and blue circles, PR-ML fitted on libraries 1--5 in setting~1; open circles, measured current at
1.70~V$_\mathrm{RHE}$. Predicted and measured endpoints are
drawn slightly apart along the potential axis so that both are visible. \textbf{c}, Largest measured
current density in each of the twelve Ni--Pd--Pt--Ru libraries at 1.60, 1.70 and 1.80~V$_\mathrm{RHE}$;
dashed line, 10~\si{\milliamperepercmsquared}. \textbf{d}, Cumulative fraction of the 251 DASH
catalysts by the largest measured current density of their curve; dashed line,
50~\si{\milliamperepercmsquared}; grey horizontal line, fraction of catalysts below
50~\si{\milliamperepercmsquared}. The measured DASH curves end near 51~\si{\milliamperepercmsquared},
where the recorded current stops increasing (\hyperref[sec:methods]{Methods}). \textbf{e}, PR-ML: the Tafel term extends the value
measured at the highest measured potential to the endpoint with a slope
estimated from the measured curves, and the residual term is added to it. Setting~1, fitted on
other libraries; setting~2, fitted on the endpoints of a few catalysts of the new library.}
\label{fig:overview}
\end{figure}

In the Ni--Pd--Pt--Ru libraries, no measurement area reached 10~\si{\milliamperepercmsquared}
within the measured window. The highest current densities at 1.60, 1.70 and 1.80~V$_\mathrm{RHE}$ were 4.98,
6.90 and 8.82~\si{\milliamperepercmsquared} (Fig.~\ref{fig:overview}c). In DASH, 79 of the 251 measured catalysts
(31\%) did not reach 50~\si{\milliamperepercmsquared} (Fig.~\ref{fig:overview}d). The endpoints used to compare OER
catalysts therefore lie beyond the measured part of many of these curves.


To test predictions against measured values, we placed the endpoint inside the measured window. For every
Ni--Pd--Pt--Ru catalyst, the currents at 1.40 and 1.55~V$_\mathrm{RHE}$ were the measured values
given to the models, and the current at 1.70~V$_\mathrm{RHE}$ was the endpoint (Fig.~\ref{fig:overview}b). PR-ML was fitted in two settings (Fig.~\ref{fig:overview}e). In setting~1, it was fitted on other libraries, and no
endpoint of the new library was used. In setting~2, it was fitted on the endpoints of 3 to 40
catalysts measured in the new library, chosen either at random or only from the less active half
of the library (range-shifted selection), with 50 repetitions for each number of labelled
catalysts.

We used the Ni--Ru library (library~6) to test prediction above the training range. Its largest
current at 1.70~V$_\mathrm{RHE}$ was 6.90~\si{\milliamperepercmsquared}, against
1.66~\si{\milliamperepercmsquared} in libraries 1--5, and 282 of its 322 catalysts exceeded the
largest training current. For 144 of them, the logarithm of the current exceeded the training
maximum by more than two interquartile ranges (IQR) of the logarithmic training currents; we call
these the catalysts far above the training range. Libraries 7--12 (2,019 catalysts), predicted
after fitting on libraries 1--6, lie inside the training range.

\subsection*{Without endpoint measurements in the new library, PR-ML predicts currents above the training range}

\begin{figure}[!tbp]
\centering
\includegraphics[width=\textwidth]{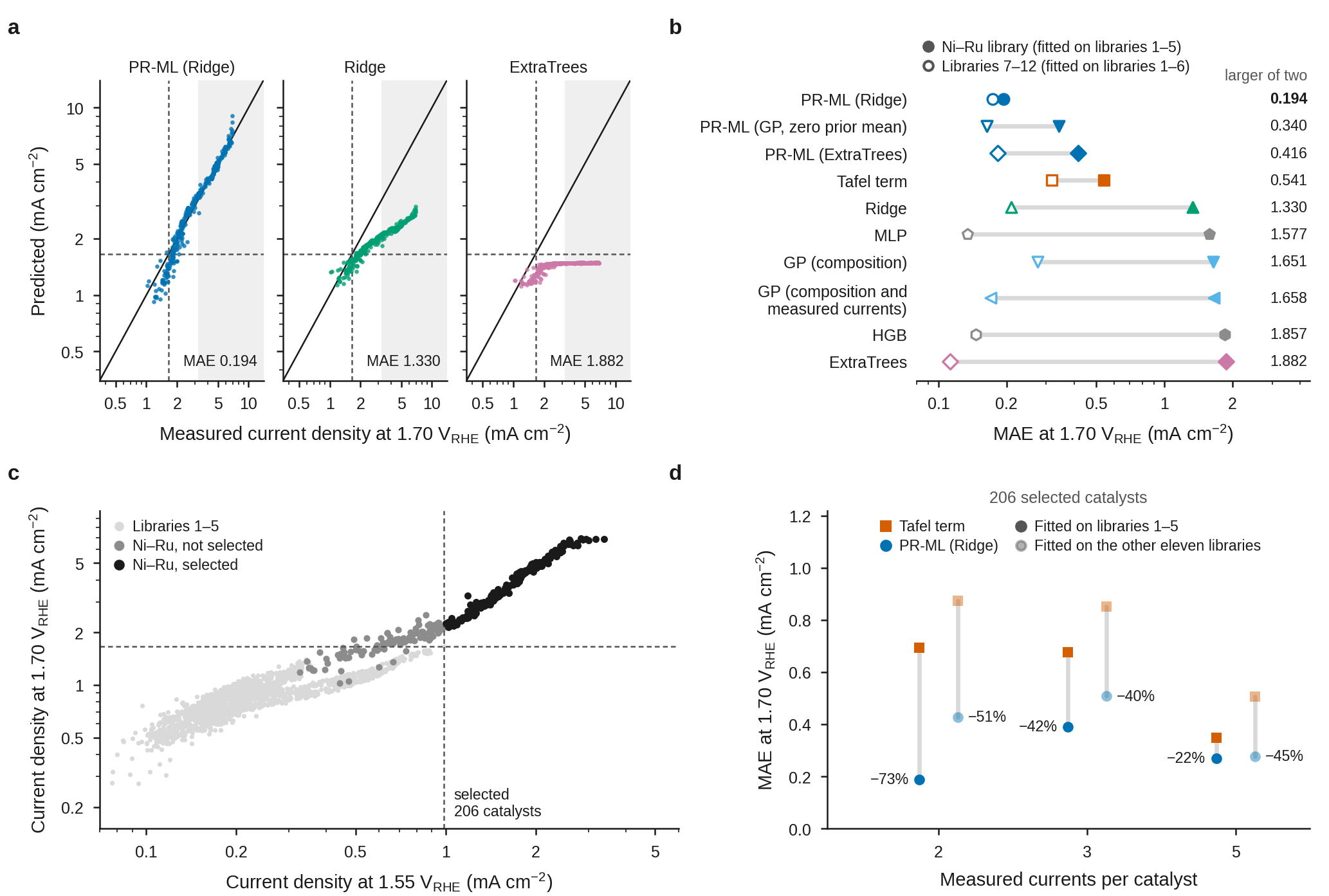}
\caption{\textbf{Setting 1: prediction without endpoint measurements in the new library.}
\textbf{a}, Predicted against measured current density at 1.70~V$_\mathrm{RHE}$ for the 322
catalysts of the Ni--Ru library, for PR-ML (Ridge), Ridge and ExtraTrees fitted on libraries 1--5 with
two measured currents per catalyst. Solid line, equal values; dashed lines, largest training
current (1.66~\si{\milliamperepercmsquared}); grey region, measured currents of the catalysts far
above the training range; numbers, mean absolute error (MAE) in \si{\milliamperepercmsquared}.
\textbf{b}, MAE of each model in the Ni--Ru library (filled; fitted on libraries 1--5) and in
libraries 7--12 (open; fitted on libraries 1--6), for the models of Table~\ref{tab:setting1} except the two
calibrations. The number on the right is the larger of the two,
which is the Ni--Ru value for every model. \textbf{c}, Current density at 1.70~V$_\mathrm{RHE}$
against current density at 1.55~V$_\mathrm{RHE}$ for the catalysts of libraries 1--5 (light grey)
and of the Ni--Ru library; dashed lines, the largest values of libraries 1--5 at the two potentials.
Ni--Ru catalysts to the right of the vertical dashed line (black points) are selected before any
endpoint is measured; mid-grey points, Ni--Ru catalysts not selected. \textbf{d}, MAE of the Tafel term and of PR-ML (Ridge) for the 206 selected catalysts, with two, three
and five measured currents per catalyst, fitted on libraries 1--5 (dark) or on the other eleven libraries
(light); squares, Tafel term; circles, PR-ML (Ridge); numbers, relative change in MAE from the Tafel term to PR-ML (Ridge).}
\label{fig:setting1}
\end{figure}

In setting~1, PR-ML was fitted on libraries 1--5 and given two measured currents per Ni--Ru
catalyst. The slope of the Tafel term was estimated from the measured curves of the training catalysts, recorded
at 17 potentials between 1.40 and 1.55~V$_\mathrm{RHE}$, and from the catalyst's own two measured
currents. The residual learner was Ridge, set up in the same way as Ridge used alone and fitted on the same
features and training catalysts, so that PR-ML (Ridge) and Ridge are compared on equal terms.
PR-ML and the selection of its settings are described in \hyperref[sec:methods]{Methods}. PR-ML (Ridge)
predicted the 1.70~V$_\mathrm{RHE}$ current of the 322 Ni--Ru catalysts with an MAE of 0.194~\si{\milliamperepercmsquared} (Table~\ref{tab:setting1}, Fig.~\ref{fig:setting1}a). The data-driven
models, namely Ridge, ExtraTrees, histogram gradient boosting (HGB), a multilayer
perceptron (MLP) and two GPs, reached 1.330--1.882~\si{\milliamperepercmsquared}. For the
catalysts far above the training range, the MAE of PR-ML (Ridge) was 0.177~\si{\milliamperepercmsquared},
against 2.486--3.363~\si{\milliamperepercmsquared} for the data-driven models.

\begin{table}[!tbp]
\centering
\caption{\textbf{Setting 1: prediction of the current density at 1.70~V$_\mathrm{RHE}$ without
endpoint measurements in the new library.} MAE in \si{\milliamperepercmsquared}. Ni--Ru library:
models fitted on libraries 1--5; ``far above'' denotes the 144 catalysts whose logarithmic current
exceeds the training maximum by more than two IQR of the logarithmic training currents, and $\rho$
is their Spearman correlation. Libraries 7--12: models fitted on libraries 1--6 (all catalysts
inside the training range). Larger of two: the larger of the Ni--Ru and libraries 7--12 MAEs. Each predicted catalyst is given two measured currents (1.40 and 1.55~V$_\mathrm{RHE}$).
PR-ML (Ridge), PR-ML (GP, zero prior mean) and PR-ML (ExtraTrees): PR-ML with Ridge, a GP whose prior
mean is zero, or ExtraTrees as the residual learner; each residual learner is set up in the same way as
the same learner used alone (\hyperref[sec:methods]{Methods}). MAE, mean absolute error; GP, Gaussian process; MLP, multilayer perceptron; HGB, histogram
gradient boosting. Calibrations were fitted on libraries 1--5 only; --, not computed.}
\label{tab:setting1}
\footnotesize
\setlength{\tabcolsep}{4pt}
\begin{tabular}{lccccc}
\toprule
 & \multicolumn{3}{c}{Ni--Ru library} & Libraries 7--12 & Larger \\
\cmidrule(lr){2-4}
Model & All (322) & Far above (144) & $\rho$, far above & All (2,019) & of two \\
\midrule
PR-ML (Ridge) & \textbf{0.194} & \textbf{0.177} & 0.98 & 0.173 & \textbf{0.194} \\
PR-ML (GP, zero prior mean) & 0.340 & 0.518 & 0.98 & 0.163 & 0.340 \\
PR-ML (ExtraTrees) & 0.416 & 0.565 & 0.98 & 0.182 & 0.416 \\
Tafel term & 0.541 & 0.735 & 0.98 & 0.318 & 0.541 \\
Tafel term with isotonic calibration & 0.531 & 0.882 & 0.98 & -- & -- \\
Tafel term with affine calibration & 1.298 & 2.328 & 0.98 & -- & -- \\
Ridge & 1.330 & 2.486 & 0.97 & 0.209 & 1.330 \\
MLP & 1.577 & 2.786 & 0.95 & 0.134 & 1.577 \\
GP (composition) & 1.651 & 2.954 & $-0.31$ & 0.275 & 1.651 \\
GP (composition and measured currents) & 1.658 & 3.147 & $-0.97$ & 0.170 & 1.658 \\
HGB & 1.857 & 3.337 & $-0.14$ & 0.147 & 1.857 \\
ExtraTrees & 1.882 & 3.363 & 0.30 & 0.112 & 1.882 \\
\bottomrule
\end{tabular}
\end{table}

The data-driven models predicted currents below the measured values for these catalysts. Ridge,
ExtraTrees, HGB and MLP predicted a lower current than the measured one for all 144
catalysts far above the training range, with median ratios of predicted to measured current of
0.31--0.49. The largest predictions of ExtraTrees and HGB, 1.49 and
1.54~\si{\milliamperepercmsquared}, were below the largest training current of
1.66~\si{\milliamperepercmsquared}. For PR-ML (Ridge), the median ratio was 1.00. For the order of these
catalysts, the Spearman correlation was $-0.31$ for GP (composition), $-0.97$ for GP (composition and
measured currents), $-0.14$ for HGB and 0.30 for ExtraTrees, against 0.95--0.98 for Ridge, MLP, PR-ML (Ridge) and the
Tafel term (Table~\ref{tab:setting1}).

Both terms of PR-ML contribute to the prediction. The Tafel term gave an MAE of
0.541~\si{\milliamperepercmsquared} (0.735 for the catalysts far above the training range), and
the sum of the Tafel term and the residual term gave 0.194 (0.177). Calibrations that apply one
function to the Tafel term of every catalyst gave 1.298 (affine) and
0.531~\si{\milliamperepercmsquared} (isotonic). With residual learners other than Ridge, PR-ML (GP, zero prior mean)
and \mbox{PR-ML} (ExtraTrees) gave 0.340 (0.518) and 0.416 (0.565)~\si{\milliamperepercmsquared}, against 1.658 (3.147)
for GP (composition and measured currents) and 1.882 (3.363) for ExtraTrees used alone; both errors were lower than those of every data-driven model.

Inside the training range, in libraries 7--12, ExtraTrees had the lowest MAE
(0.112~\si{\milliamperepercmsquared}), the Tafel term the highest (0.318), and PR-ML (Ridge) 0.173~\si{\milliamperepercmsquared} (Table~\ref{tab:setting1}). Whether a new library lies above the
training range is not known before its endpoints are measured, so we compared the larger of each
model's two errors. For PR-ML (Ridge) this value was 0.194~\si{\milliamperepercmsquared}, the lowest of
all compared models; it was 0.340 for PR-ML (GP, zero prior mean), 0.416 for PR-ML (ExtraTrees), 0.541
for the Tafel term and 1.330--1.882~\si{\milliamperepercmsquared} for the
data-driven models (Fig.~\ref{fig:setting1}b).

For the 144 catalysts far above the training range, PR-ML (Ridge) and PR-ML (ExtraTrees) had a lower
error than the same learner used alone with each of 1, 2, 3, 5, 9
and 17 measured currents. This advantage held in four further tests. First, neighbouring measurement areas have similar compositions, so their errors may be
correlated; we therefore resampled whole clusters of catalysts with similar compositions instead of
single catalysts (difference in MAE between PR-ML (Ridge) and Ridge with two measured currents: $-2.31$~\si{\milliamperepercmsquared}, 95\% confidence interval $-2.55$ to
$-2.00$ with 12 clusters; Supplementary Table~\ref{tab:si_cluster}). Second, the slope of the Tafel term
was estimated only from the currents of the training catalysts at the measured potentials; the advantage
held with 2, 3, 5, 9 and 17 measured currents and for every group of catalysts, and
the currents at 17 potentials, which are already recorded for the training catalysts, lowered the error
further with two, three and five measured currents (Supplementary Table~\ref{tab:si_slope}). Third, the
advantage held for all 60 sets of two, three, five or nine measured potentials that we tested. Fourth,
normally distributed errors with a standard deviation of up to 0.10 in $\log_{10}$ current were added
to the measured currents; with two, three and five measured currents, the advantage held in at least 10
of 12 repetitions at every standard deviation. With nine and seventeen measured currents,
neighbouring measured potentials lie only 19 and 9~mV apart; when the catalyst's own slope was fitted
over all measured potentials instead of taken from the two highest, the advantage held in all 12
repetitions at every standard deviation. With closely spaced measured potentials, the catalyst's own slope should therefore be fitted over all measured currents (Supplementary Table~\ref{tab:si_robust}).

The advantage of PR-ML over the Tafel term was largest when two or three currents were measured per
catalyst at
evenly spaced potentials: with two measured currents, the MAE for the catalysts far above the training
range was 0.177~\si{\milliamperepercmsquared} for PR-ML (Ridge) and 0.735 for the Tafel term. With nine or
seventeen measured currents (0.179 and 0.172~\si{\milliamperepercmsquared}) or with the measured
potentials at the high end of the window (0.172~\si{\milliamperepercmsquared}), the Tafel term, which
uses no endpoint labels, was as accurate as PR-ML or more accurate for these catalysts, so for catalysts far above the training range these measurement
designs need no endpoint labels (Supplementary Tables~\ref{tab:si_robust}a and~\ref{tab:si_budget}).

PR-ML (Ridge) also remained accurate with few training catalysts. In each of 20 random subsets, we fitted
it on 35 of the 1,685 catalysts of libraries 1--5 (seven per library), with two measured currents. For
the catalysts far above the training range, its median MAE over the 20 random subsets was 0.228~\si{\milliamperepercmsquared}, against
2.686 for Ridge fitted on the same 35 catalysts and 2.486 for Ridge fitted on all 1,685 (Supplementary Table~\ref{tab:si_reduction}).

\subsection*{Catalysts above the training range can be selected before any endpoint is measured}

Libraries that contain catalysts above the training range can be recognized from the last measured
current. We
selected the catalysts whose current at 1.55~V$_\mathrm{RHE}$ exceeded the highest
1.55~V$_\mathrm{RHE}$ current of the training libraries. When each library was predicted from the
other eleven, only the Ni--Ru library contained such catalysts, and it was also the only library
with catalysts above the training range (Supplementary Table~\ref{tab:si_selection}). With libraries 1--5 as training
libraries (Fig.~\ref{fig:setting1}c), the rule selected 206 catalysts, all of which lay above the training range, and did not select 76 of
the 282 catalysts above it; with the other eleven libraries, it selected 206 catalysts, of which
185 lay above the training range, and did not select 1 of the 186 above it.

For the 206 selected catalysts, the Tafel term was already more accurate than Ridge and ExtraTrees
(Supplementary Table~\ref{tab:si_selection}), and PR-ML (Ridge) was more accurate still. With two, three and five currents measured between 1.40
and 1.55~V$_\mathrm{RHE}$, the MAE fell from 0.695 to 0.188, from 0.676 to 0.390 and from 0.348 to
0.271~\si{\milliamperepercmsquared} (reductions of 73\%, 42\% and 22\%) when the model was fitted
on libraries 1--5, and by 51\%, 40\% and 45\% when it was fitted on the other eleven libraries
(Fig.~\ref{fig:setting1}d).

\subsection*{In the Ni--Pd--Pt--Ru libraries, standard Tafel slopes give large endpoint errors}

\begin{figure}[!tbp]
\centering
\includegraphics[width=\textwidth]{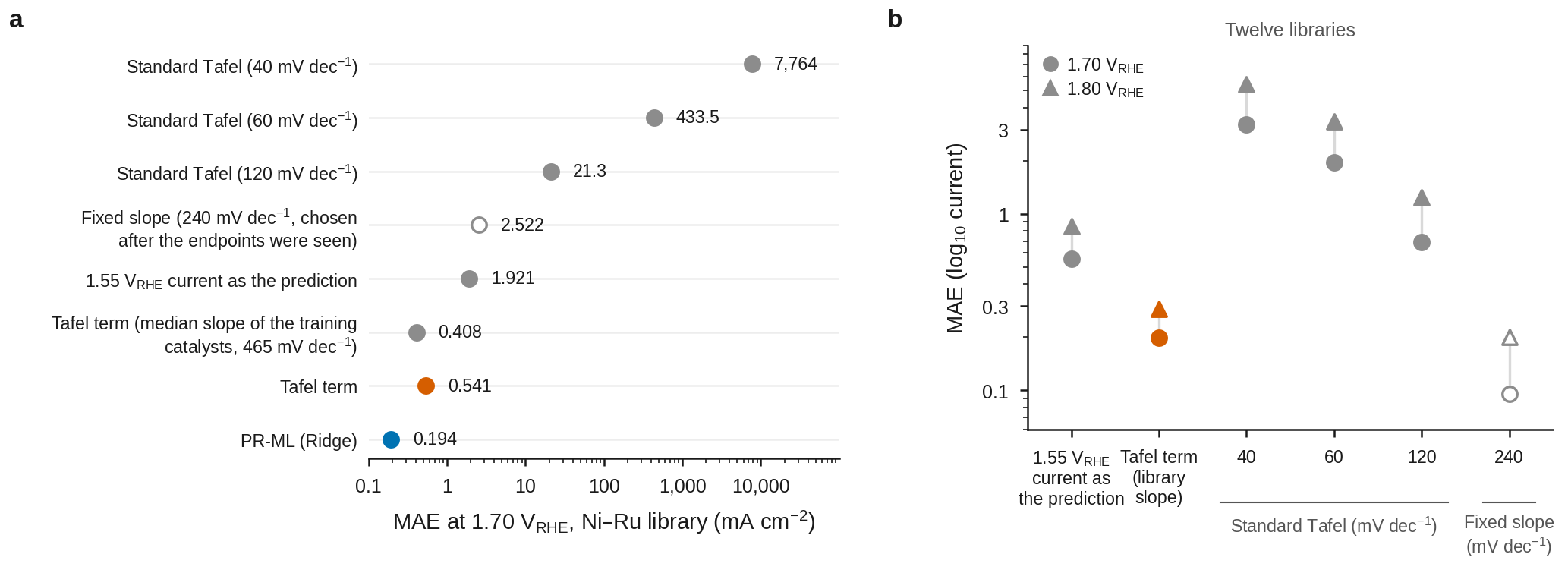}
\caption{\textbf{Standard Tafel slopes give large endpoint errors in the Ni--Pd--Pt--Ru libraries.}
\textbf{a}, Mean absolute error (MAE) at 1.70~V$_\mathrm{RHE}$ in the Ni--Ru library, with no
endpoint of this library used (logarithmic axis). Grey, standard Tafel, the 1.55~V$_\mathrm{RHE}$ current as the prediction and the Tafel term
(median slope of the training catalysts); open, the fixed slope (240~mV~dec$^{-1}$, chosen after the
endpoints were seen); vermillion, the Tafel term; blue, PR-ML (Ridge). \textbf{b}, MAE in
$\log_{10}$ current over the twelve libraries at 1.70~V$_\mathrm{RHE}$ (circles) and
1.80~V$_\mathrm{RHE}$ (triangles) for the 1.55~V$_\mathrm{RHE}$ current as the prediction, the Tafel
term (library slope), standard Tafel and the fixed slope. Vermillion, the Tafel term (library slope),
which is the Tafel term of setting~2; its slope is the median of the two-point slopes of all catalysts
in each library. No endpoint
labels are used, except that the fixed slope of 240~mV~dec$^{-1}$ (open symbols) was chosen after the
endpoints were seen.}
\label{fig:slope}
\end{figure}

Between 1.40 and 1.55~V$_\mathrm{RHE}$, the median slope of the training catalysts was
465~mV~dec$^{-1}$, far above the standard Tafel slopes, and the Tafel equation with a standard slope (standard
Tafel) gave large endpoint errors
(Fig.~\ref{fig:slope}). In the Ni--Ru library, with no endpoint of this library used, standard Tafel (40,
60 and 120~mV~dec$^{-1}$) gave MAEs of 7,764, 433.5 and 21.3~\si{\milliamperepercmsquared}, and using the
1.55~V$_\mathrm{RHE}$ current as the prediction gave 1.921~\si{\milliamperepercmsquared}. With slopes estimated from the measured curves, the Tafel term (median slope of the training catalysts)
gave 0.408~\si{\milliamperepercmsquared}, and the Tafel term, whose slope also uses the catalyst's own
two measured currents, gave 0.541~\si{\milliamperepercmsquared}; PR-ML (Ridge) reduced the error to
0.194~\si{\milliamperepercmsquared}.

Over the twelve libraries, the Tafel term (library slope), whose slope is the median of the two-point
slopes of all catalysts in the library and which uses no endpoint labels, gave MAEs of 0.200 and 0.289 in $\log_{10}$ current
at 1.70 and 1.80~V$_\mathrm{RHE}$. Standard Tafel (40, 60 and 120~mV~dec$^{-1}$) gave 0.694--3.194 and 1.235--5.401, which
correspond to currents that differ from the measured ones by a factor of about 5 to 1,563 at
1.70~V$_\mathrm{RHE}$ (geometric mean over catalysts; Fig.~\ref{fig:slope}b). The fixed slope (240~mV~dec$^{-1}$, chosen after the endpoints had been seen) gave 0.095 in $\log_{10}$
current over the twelve libraries but 2.522~\si{\milliamperepercmsquared} in the Ni--Ru library, so
neither standard Tafel nor this fixed slope gave low errors in every library.

\subsection*{With a few endpoints from the new library, PR-ML reaches the accuracy obtained from other libraries}

\begin{figure}[!tbp]
\centering
\includegraphics[width=\textwidth]{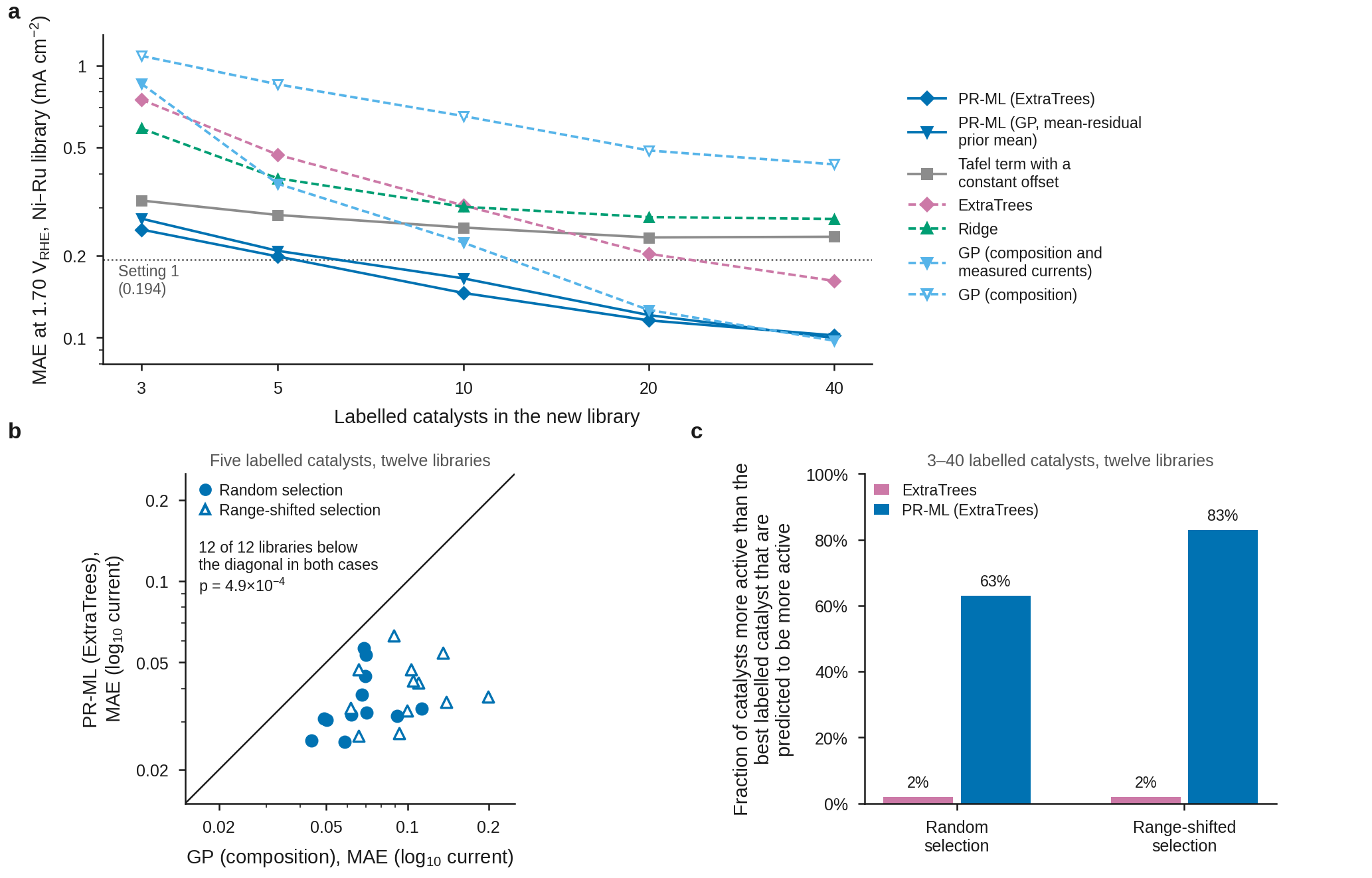}
\caption{\textbf{Setting 2: prediction with a few endpoints measured in the new library.}
\textbf{a}, Mean absolute error (MAE) at 1.70~V$_\mathrm{RHE}$ in the Ni--Ru library against the
number of labelled catalysts chosen at random (mean of 50 repetitions). Solid blue line and
diamonds, PR-ML (ExtraTrees); solid blue line and triangles, PR-ML (GP, mean-residual prior mean), whose
GP prior mean is the mean labelled residual; grey solid line and squares, the Tafel term with a constant offset, which
in setting~2 equals the 1.55~V$_\mathrm{RHE}$ current multiplied by a factor fitted on the labelled
catalysts; dashed lines, ExtraTrees, Ridge, GP (composition) and GP (composition and measured currents).
Dotted line, the setting~1 error of PR-ML (Ridge), 0.194~\si{\milliamperepercmsquared}. \textbf{b}, MAE in $\log_{10}$ current of each of the twelve
libraries with five labelled catalysts, PR-ML (ExtraTrees) against GP (composition), for random (filled circles) and range-shifted (open triangles)
selection; points below the diagonal are libraries where PR-ML (ExtraTrees) is more accurate ($p$, Wilcoxon
signed-rank test over libraries). \textbf{c}, Fraction of the catalysts more active than the best
labelled catalyst that are predicted to be more active than it, averaged over the twelve libraries,
3--40 labelled catalysts and 50 repetitions (a repetition with no such catalyst counts as zero),
for ExtraTrees and PR-ML (ExtraTrees). GP, Gaussian process.}
\label{fig:setting2}
\end{figure}

In setting~2, PR-ML (ExtraTrees) predicted the 1.70~V$_\mathrm{RHE}$
current of the Ni--Ru library with MAEs of 0.250, 0.199, 0.146, 0.116 and
0.102~\si{\milliamperepercmsquared} with 3, 5, 10, 20 and 40 labelled catalysts chosen at random
(Fig.~\ref{fig:setting2}a). With five labelled catalysts, its MAE of 0.199~\si{\milliamperepercmsquared} was about the same as
that of PR-ML (Ridge) in setting~1 (0.194~\si{\milliamperepercmsquared}). ExtraTrees reached
0.203~\si{\milliamperepercmsquared} only with 20 labelled catalysts, Ridge stayed at
0.273~\si{\milliamperepercmsquared} with 40, and GP (composition), of the type used in the original study,
reached 0.853 with five and 0.434~\si{\milliamperepercmsquared} with 40. GP (composition and measured
currents) reached 0.368 with five and 0.097~\si{\milliamperepercmsquared} with 40 labelled
catalysts.

Over all twelve libraries, PR-ML (ExtraTrees) had the lowest error with five labelled catalysts
(Supplementary Table~\ref{tab:si_twelve}). With five labelled catalysts, its MAE in $\log_{10}$ current was 0.036 against
0.048--0.068 for the data-driven models with random selection, and 0.041 against 0.078--0.106 with
range-shifted selection. It was more accurate than GP (composition) in all twelve libraries with both
selections (Fig.~\ref{fig:setting2}b; Wilcoxon signed-rank test over libraries, $p=4.9\times10^{-4}$), and the difference
doubled from 0.032 with random selection to 0.065 with range-shifted selection. In all four
combinations of selection and number of labels (5 and 20), the lowest error came from PR-ML with ExtraTrees or
a GP as the residual learner (Supplementary Table~\ref{tab:si_twelve}).

PR-ML (ExtraTrees) also predicted which catalysts exceed the best labelled catalyst. Averaged over the
twelve libraries and 3--40 labelled catalysts, it predicted 63\% of these catalysts to exceed the
best labelled catalyst with random selection and 83\% with range-shifted selection, against 2\% for
ExtraTrees (Fig.~\ref{fig:setting2}c). When the measurement stopped at 1.55~V$_\mathrm{RHE}$ and the
endpoint was 1.70~V$_\mathrm{RHE}$, PR-ML was more accurate than the most accurate compared method,
or showed no significant difference from it, in all 20 conditions (two pairs of measured potentials
ending at 1.55~V$_\mathrm{RHE}$, two selections and 3--40 labelled catalysts), for catalysts below and
above the largest label and for all catalysts (Supplementary Table~\ref{tab:si_stop155}).

\subsection*{PR-ML works with different residual learners, also in two independent datasets}

With endpoints from the new library, the residual learner can be exchanged. With five labelled catalysts in the Ni--Ru library, PR-ML (GP, mean-residual prior mean), whose GP prior
mean is the mean labelled residual, gave an MAE of 0.208~\si{\milliamperepercmsquared}, against 0.199 for
PR-ML (ExtraTrees). Over the twelve libraries, PR-ML (ExtraTrees) gave the lower error in 11 (random selection) and 10
(range-shifted selection) of the 12 libraries, by 0.0017 and 0.0032 in $\log_{10}$ current (0.4\%
and 0.7\% of the current; $p=0.034$).

\begin{figure}[!tbp]
\centering
\includegraphics[width=\textwidth]{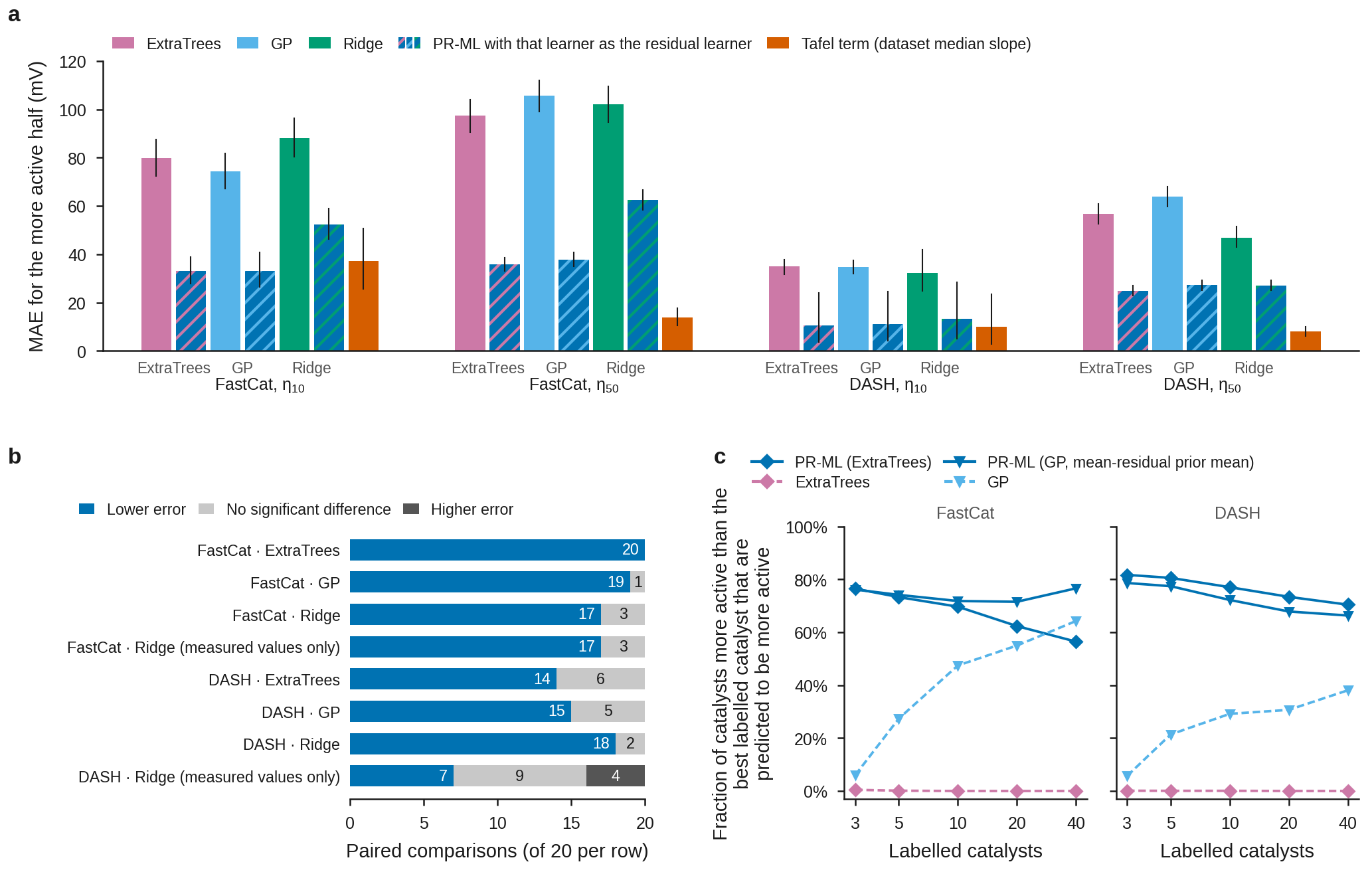}
\caption{\textbf{Prediction above the range of the labels in two independent datasets.}
\textbf{a}, Mean absolute error (MAE) of the overpotential for the more active half of the FastCat
and DASH catalysts,
with five labelled catalysts from the less active half. $\eta_{10}$ and $\eta_{50}$ are the
overpotentials at 10 and 50~\si{\milliamperepercmsquared}; $\eta_{10}$ is predicted from the
overpotentials at 1, 2 and 5~\si{\milliamperepercmsquared} ($\eta_{10}$ design) and $\eta_{50}$
from those at 5 and 10~\si{\milliamperepercmsquared} ($\eta_{50}$ design). In each group, the pink, sky-blue and green
bars are ExtraTrees, GP and Ridge, each followed by a blue bar, hatched in the colour of that learner,
for PR-ML with that learner as the residual learner; the last bar, in vermillion, is the Tafel term (dataset median slope), which uses no
endpoint labels. Error bars, 95\% confidence intervals from
2,000 bootstrap resamples of the catalysts. \textbf{b}, Paired comparisons of PR-ML with each learner as the residual learner and the same learner
used alone, over the two designs, two label fractions (50\% and 70\%) and 3--40 labelled catalysts:
number of comparisons in which the error of PR-ML was lower, not significantly different or higher.
Ridge uses composition and measured overpotentials, and Ridge (measured values only) uses the measured
overpotentials only. \textbf{c}, Fraction of
the catalysts more active than the best labelled catalyst that are predicted to be more active than
it, averaged over the two designs and the two label fractions. Solid blue line and diamonds, PR-ML (ExtraTrees); solid blue
line and triangles, PR-ML (GP, mean-residual prior mean); dashed lines, ExtraTrees and GP.
GP, Gaussian process.}
\label{fig:independent}
\end{figure}

In FastCat and DASH, we tested prediction above the label range with the endpoint defined
as the overpotential at 10 or 50~\si{\milliamperepercmsquared} ($\eta_{10}$ or $\eta_{50}$;
Fig.~\ref{fig:independent}). Catalysts were sorted by their measured endpoint, labelled catalysts were drawn only from
the less active 50\% or 70\% (the label fraction), and the remaining, more active catalysts were
predicted from their overpotentials at lower current densities. The endpoints of the more active
catalysts were lower than the best endpoint of the less active 50\% by a median of 19--45~mV and by
up to 114~mV (9--23~mV and up to 87~mV for the less active 70\%). Each comparison was paired: one model used a learner alone, and PR-ML used the same learner, set up in
the same way and fitted on the same features and labels, as the residual learner. We report two measurement designs: $\eta_{10}$ predicted from the
overpotentials at 1, 2 and 5~\si{\milliamperepercmsquared}, and $\eta_{50}$ from those at 5 and
10~\si{\milliamperepercmsquared}.

With five labelled catalysts and a label fraction of 50\%, ExtraTrees predicted the
more active catalysts with MAEs of 79.9, 97.4, 34.9 and 56.7~mV (FastCat $\eta_{10}$, FastCat
$\eta_{50}$, DASH $\eta_{10}$, DASH $\eta_{50}$), and PR-ML (ExtraTrees) with 33.0, 35.8, 10.5 and
25.0~mV (Fig.~\ref{fig:independent}a). The corresponding MAEs were
74.3, 105.8, 34.8 and 63.9~mV for GP and 33.1, 37.9, 11.1 and 27.2~mV for PR-ML (GP, mean-residual
prior mean). Over the two
datasets, two measurement designs, two label fractions and 3--40 labelled catalysts, the two PR-ML models had a lower error
than the same learner used alone in 68 of 80 paired comparisons and showed no significant difference in the other 12. PR-ML (Ridge) uses Ridge on composition
and measured overpotentials as the residual learner, and PR-ML (Ridge, measured values only) uses Ridge
on the measured overpotentials only. With these two included, PR-ML had a lower error than the same
learner used alone in 127 of 160 comparisons, showed no significant difference in 29 and had a higher error in 4, by at most 2.6~mV
(Fig.~\ref{fig:independent}b). Over all 16 combinations of measurement design and endpoint, PR-ML with ExtraTrees or with either
of the two Ridge learners as the residual learner had a lower error than the same learner used alone in
624 of 960 comparisons, showed no significant difference in 237 and had a higher error in 99; for
PR-ML (ExtraTrees), the numbers were 222, 89 and 9 of 320 (Supplementary Table~\ref{tab:si_combinations}). Used alone, ExtraTrees predicted at most 1\% of the catalysts more active than the best
labelled catalyst to be more active than it, and GP 3--91\%; with these learners as the residual learner,
PR-ML predicted 33--95\% and 38--94\%, respectively (ranges over the individual conditions; Fig.~\ref{fig:independent}c shows the averages over the two
measurement designs and two label fractions).

For $\eta_{50}$ predicted from the overpotentials at 5 and 10~\si{\milliamperepercmsquared}, no endpoint
labels were needed. The Tafel term (dataset median slope), which uses none, gave 13.9~mV (FastCat) and 8.0~mV (DASH) for the
more active catalysts with a label fraction of 50\% (Fig.~\ref{fig:independent}a). This error was lower
than that of PR-ML or of the Tafel term with a constant offset, both fitted on the less active catalysts,
in 95 of 100 comparisons (two datasets, four residual learners and the constant offset, two label
fractions and 3--40 labelled catalysts), and there was no significant difference in the other 5.
Standard Tafel matched to each dataset gave the same errors (80~mV~dec$^{-1}$ for FastCat, 13.8~mV;
60~mV~dec$^{-1}$ for DASH, 7.9~mV), and the Tafel term reached them without choosing a slope in advance.
Over the 16 combinations of design and endpoint, the Tafel term was more accurate than PR-ML
with ExtraTrees or with either of the two Ridge learners as the residual learner in 196 of 480 comparisons in
FastCat and in 438 of 480 in DASH, and less accurate in 98 and 10
(Supplementary Table~\ref{tab:si_combinations}).

In a campaign, the labelled catalysts are those made and measured earlier. We therefore also divided
the catalysts by the order in which they were made (FastCat: the order of the experiments; DASH: the
synthesis batches) and drew the labelled catalysts from the earlier ones. Over the same four residual learners, the two
measurement designs reported above, both datasets and five numbers of labelled catalysts (80
comparisons), PR-ML had a lower error than the same learner used alone in 43 of 80
comparisons for all later catalysts, showed no significant difference in 36 and had a higher error in
1; for the later catalysts more active than the best labelled catalyst, the numbers were 50, 28 and 2.
Over all 16 combinations of design and endpoint without the GP, the numbers for all later catalysts
were 213, 202 and 65 of 480. In DASH, the Tafel term (dataset median slope) was more accurate than
PR-ML in 28 of the 40 comparisons for all later
catalysts and showed no significant difference in the other 12 (8.4 and 16.3~mV for $\eta_{10}$ and
$\eta_{50}$ with five labelled catalysts); in FastCat, \mbox{PR-ML} was more
accurate than the Tafel term in 20 of 40 comparisons and less accurate in 4
(Supplementary Table~\ref{tab:si_chrono}).


\section*{Discussion}\phantomsection\label{sec:discussion}

The results lead to three insights. First, the measured part of each catalyst's own polarization
curve provides a physically based reference value for predicting endpoints above the training range.
A data-driven model learns a relation between features and endpoints from labelled catalysts. To
predict catalysts above the training range, such a model must extrapolate beyond the labelled
catalysts in feature space, and tree ensembles cannot predict beyond the largest label
\cite{ExtraTrees2006,KnownUnknowns2025}. Accordingly, in the Ni--Ru library, the data-driven models
predicted currents below the measured values for the catalysts far above the training range. The value
measured at the highest potential of a catalyst's curve, in contrast, already reflects the activity of
that catalyst, including activity above the training range. PR-ML extends this value along the catalyst's own polarization curve over a short potential interval, and for the catalysts far above the
training range, PR-ML (Ridge) gave a median ratio of predicted to measured current of 1.00. The same measured value
also identifies catalysts above the training range before any endpoint is measured: all 206 catalysts
whose current at 1.55~V$_\mathrm{RHE}$ exceeded the highest 1.55~V$_\mathrm{RHE}$ current of
libraries 1--5 lay above the training range. This use of the catalyst's own measurement distinguishes PR-ML from earlier hybrid models and from GPs with a physical mean function. In those models, the
physical term is computed from composition, from other calculated properties, from lower-level
calculations or from the input variables of the search
\cite{Mamun2020,Ramakrishnan2015DeltaML,Noack2021GP,Ziatdinov2022sGP}, and the Butler--Volmer
coefficients that Thelen et al.\ interpolate across compositions are fitted to fully measured curves
\cite{Thelen2026SECCM}.

Second, the two terms of PR-ML have different functions within one prediction: the Tafel term
extends the prediction beyond the training range, and the residual term corrects the error of the
Tafel term for each catalyst. The Tafel term extends the value measured at the catalyst's highest
potential or current density to the endpoint with a slope estimated from the measured curves. Standard Tafel gave large endpoint errors in the Ni--Pd--Pt--Ru libraries, because the slope between
1.40 and 1.55~V$_\mathrm{RHE}$ lies far above the standard Tafel slopes and changes with overpotential
\cite{Shinagawa2015,Anantharaj2021TafelPitfalls}. The remaining error of the Tafel term differs between catalysts, so one calibration function
applied to every catalyst cannot remove it, whereas the residual term of PR-ML (Ridge), predicted from the features
of each catalyst, reduced the error from 0.541 to 0.194~\si{\milliamperepercmsquared} in the Ni--Ru
library. This error was the lowest of all compared models, as was the larger of the two errors of PR-ML
(Ridge) inside and above the training range (Table~\ref{tab:setting1}). Because the residual learner fits this correction and not
the endpoint itself, it does not have to reproduce activities above the range of its labels. This
explains why few endpoint labels are enough: with five labelled catalysts from the Ni--Ru library,
PR-ML (ExtraTrees) reached the accuracy obtained from other libraries, whereas ExtraTrees needed 20; in setting~1,
PR-ML (Ridge) fitted on 35 training catalysts was more accurate than Ridge fitted on all
1,685. It also explains why learners that cannot predict beyond their largest label,
such as ExtraTrees, can serve as the residual learner: in two independent datasets, PR-ML with
ExtraTrees or a GP as the residual learner gave a lower error for the catalysts above the label range than the same learner used
alone in 68 of 80 comparisons and a higher error in none. When the Tafel term is already accurate, no
endpoint labels are needed. This was the case with nine and seventeen
measured currents for the catalysts far above the training range, and for $\eta_{50}$ predicted from
the overpotentials at 5 and 10~\si{\milliamperepercmsquared} in FastCat and DASH, where the Tafel term, which uses no endpoint labels and no slope chosen in advance, reached the
errors of standard Tafel matched to each dataset and was more accurate than PR-ML or the Tafel term
with a constant offset in 95 of 100 comparisons (Supplementary Tables~\ref{tab:si_budget}
and~\ref{tab:si_combinations}). This agrees with the condition under which the Tafel equation gives
correct endpoints, a nearly constant slope between the measured part of the curve and the endpoint.

Third, our approach adds endpoint prediction to the measurement step of a discovery campaign and
supplies these endpoints to the methods that choose which catalysts to make. Sequential learning, active learning and
Bayesian optimization choose the next catalysts from measured activities
\cite{Rohr2020Sequential,Moon2024ActiveOER,Thelen2025OER}. After the catalysts are made, each catalyst
is measured up to a potential or current density below the endpoint, and PR-ML predicts from these
measurements the endpoints that the selection methods use. In each round, PR-ML is fitted either on
earlier libraries, with no endpoint of the new library, or on about five endpoints measured in the new
library. When the labelled catalysts were those made earlier, as in a campaign, PR-ML was more accurate
than the same learner used alone, or showed no significant difference from it, in 79 of 80 comparisons
(Supplementary Table~\ref{tab:si_chrono}). A benchmarking guideline uses
1.6~V$_\mathrm{RHE}$ as the upper potential limit to avoid intense bubble formation and faster
dissolution \cite{Zlatar2023}. In this study, the currents at 1.70~V$_\mathrm{RHE}$, above this limit,
were predicted from the currents measured at 1.40 and 1.55~V$_\mathrm{RHE}$, so the catalysts being
predicted need to be measured only up to 1.55~V$_\mathrm{RHE}$. The predicted endpoints also identify
the catalysts of most interest for the next selection: over the twelve libraries, PR-ML (ExtraTrees)
identified 63--83\% of the catalysts more active than the best labelled catalyst, against 2\% for
ExtraTrees. Other methods also predict final values from incomplete measurements. Early prediction of
learning curves \cite{Domhan2015LearningCurve} and of battery cycle life
\cite{Severson2019,Attia2020} predicts the final value of a run from its first part, and freeze-thaw
Bayesian optimization uses such predictions to decide which training runs of machine-learning models
to continue \cite{Swersky2014FreezeThaw}. Multi-fidelity learning combines many inexpensive
measurements with a few expensive ones \cite{Fare2022,SabanzaGil2025}. These methods relate early and
final values through parametric curve forms or relations learned from data. In catalyst discovery, our approach
relates the measured part of each catalyst's curve to its endpoint through the Tafel equation, which
describes how the current of an electrochemical reaction increases with potential, and a residual term
learned from endpoint measurements, so that the selection methods receive endpoints above all earlier
measurements.

The study has the following limitations. The slope of the Tafel term is estimated from the measured curves, so the model does not require the
measured part to be a region of constant Tafel slope. This slope is an empirical quantity and is not
interpreted as the Tafel slope of a reaction mechanism, because slopes taken from potentiodynamic
curves can include currents other than the OER current \cite{Anantharaj2021TafelPitfalls}. With a GP as the residual learner, the model also gives prediction intervals; these need calibration before they are used as
uncertainty estimates, for example by conformal prediction, which sets the interval width from the
errors on held-out labelled catalysts \cite{Angelopoulos2023Conformal}. Predicting the performance of a
catalyst in an electrolyser, where it is tested in a membrane electrode assembly, or its long-term
stability from three-electrode laboratory measurements requires physical relations other than the
Tafel equation, because electrolysers operate under different conditions \cite{Lazaridis2022RDEMEA}.

\section*{Methods}\phantomsection\label{sec:methods}

\subsection*{Datasets}

The Ni--Pd--Pt--Ru data are Zenodo record 14891704 (CC BY 4.0) \cite{Thelen2025OERData}, associated
with ref.~\cite{Thelen2025OER}. The 38 CSV files of the record contain, for each measurement area, the
composition measured by energy-dispersive X-ray spectroscopy, the reported activity and the linear
sweep voltammogram; their MD5 checksums were verified against the record. For each library, the
composition, activity and voltammogram were matched by measurement area. This gave 4,026 catalysts
(322--340 per library), each with currents on a common grid of 771 potentials from 1.03 to
1.80~V$_\mathrm{RHE}$. The reported activity equals the current density at 1.70~V$_\mathrm{RHE}$ of the
voltammogram. The libraries are numbered in the order in which they were prepared, and the training
libraries of each prediction are defined by these numbers.

FastCat \cite{FastCat2025,FastCatData2025} (version 1, MIT licence) contains 101 experiments of a
Bayesian optimization campaign. We kept one experiment per nominal composition, the first recorded,
which gave 89 catalysts. At each current density of 1, 2, 5, 10, 20 and
50~\si{\milliamperepercmsquared}, the potential was averaged over the last 20~s of the constant-current
step and corrected for 90\% of the ohmic drop, as in the original study. The overpotential is this
potential minus 1.23~V$_\mathrm{RHE}$. The recomputed overpotentials at
10~\si{\milliamperepercmsquared} agree with the published values to within 1.5~mV.

DASH \cite{DASH2025,DASHData2025} (CC0) contains 258 polarization curves. After removing five curves of
reference catalysts and two curves that could not be matched to a single entry of the synthesis table
of the dataset, 251 curves remained. The current density is the recorded current divided by the
electrode area of 0.196~cm$^2$. Before interpolation, the points of each curve were ordered by
potential and the curve was taken from its lowest current onward. Each current value was then replaced
by the largest current recorded up to that potential, and points without an increase in current were
removed, so that the current increases with potential. The potential at each of the six current
densities was obtained by linear interpolation, and the overpotential is this potential minus
1.23~V$_\mathrm{RHE}$.
The recorded potentials were used without further correction. With this area, the overpotentials at 10~\si{\milliamperepercmsquared} agree with the
published values to within 2~mV for 241 of the 250 curves with a published value. In the 169 curves
that reach the current limit of the instrument, the recorded current stops increasing at 10.0--10.2~mA,
that is, near 51~\si{\milliamperepercmsquared}, so 50~\si{\milliamperepercmsquared} is the highest
current density used as an endpoint. The 135 curves whose measured range includes all six current
densities were used.

\subsection*{Measured values, endpoints and features}

For the Ni--Pd--Pt--Ru libraries, the endpoint is the current density at 1.70~V$_\mathrm{RHE}$
(1.80~V$_\mathrm{RHE}$ in Fig.~\ref{fig:slope}b), and the models predict its logarithm,
$\log_{10}[j/(\si{\milliamperepercmsquared})]$. The measured values given to the models are the
logarithms of the current densities at 1.40 and 1.55~V$_\mathrm{RHE}$. With three and five measured
currents (Fig.~\ref{fig:setting1}d), the measured potentials were 1.40, 1.475 and
1.55~V$_\mathrm{RHE}$, and 1.40, 1.438, 1.475, 1.512 and 1.55~V$_\mathrm{RHE}$. The features are the
atomic fractions of Ni, Pd, Pt and Ru and the measured values.

For FastCat and DASH, a measurement design is the set of current densities at which the overpotential
is given to the models, and the endpoint is the overpotential at a higher current density. Six designs,
each combined with every endpoint ($\eta_{10}$, $\eta_{20}$, $\eta_{50}$) above its highest measured
current density, gave 16 combinations. The main text reports $\eta_{10}$ predicted from the
overpotentials at 1, 2 and 5~\si{\milliamperepercmsquared}, and $\eta_{50}$ predicted from those at 5
and 10~\si{\milliamperepercmsquared}. The features are the nominal atomic
fractions of the elements supplied by the precursors (8 elements in FastCat and 32 in DASH, where F and
N were excluded and the remaining fractions renormalized) and the measured values.

\subsection*{Physics-residual machine learning (PR-ML)}

For a catalyst $x$, $y(x)$ is its measured endpoint ($\log_{10} j$ for the Ni--Pd--Pt--Ru libraries
and the overpotential for FastCat and DASH) and $\hat{y}(x)$ its prediction. The Tafel term extends the value measured at the highest measured potential,
$E_\mathrm{m}$, to the endpoint potential $E_\mathrm{e}$ \cite{Stern1957}:
\begin{equation}
 T(x)=\log_{10} j(E_\mathrm{m})+s\,(E_\mathrm{e}-E_\mathrm{m}),
 \label{eq:tafel}
\end{equation}
where $s$ is the slope in decades of current per volt, which corresponds to $1000/s$
mV~dec$^{-1}$. When the endpoint is an overpotential (FastCat and DASH), the same form extends the
overpotential at the highest measured current density $j_\mathrm{m}$ to the endpoint current density
$j_\mathrm{e}$: $T(x)=\eta(j_\mathrm{m})+b\log_{10}(j_\mathrm{e}/j_\mathrm{m})$, where $b$ is the
slope in mV~dec$^{-1}$. The slopes are estimated from measured curves and not from endpoints, and a negative slope
in the Tafel term is set to zero.
\begin{itemize}
 \item In setting~1, $s=(1-w)\,g+w\,s_\mathrm{c}$ with $w=0.25$. To obtain $g$, the logarithm of the
 current of each training catalyst was fitted by least squares against potential at 17 evenly spaced
 potentials between 1.40 and 1.55~V$_\mathrm{RHE}$, and $g$ is the median of these slopes
 (2.15 decades per volt, or 465~mV~dec$^{-1}$, for libraries 1--5). The catalyst's own slope $s_\mathrm{c}$ is the difference in
 the logarithm of the current between its two highest measured potentials divided by their potential
 difference; with one measured current, $s=g$.
 \item In setting~2, $s$ is the library slope: the median of the two-point slopes between the measured
 potentials of all catalysts of the new library.
 \item In FastCat and DASH, the overpotential of each catalyst was fitted by least squares against
 $\log_{10} j$ at its measured current densities, and $b$ is the dataset median slope: the median of
 these slopes over all catalysts of the dataset.
\end{itemize}

The residual learner is a regression model fitted, over the fitting catalysts, to $y(x)-T(x)$, and
$r(x)$ is its prediction for catalyst $x$. The fitting catalysts are the training catalysts in
setting~1 and the labelled catalysts in setting~2 and in FastCat and DASH. We call the measured
endpoints of the labelled catalysts labels, and each random choice of labelled catalysts a draw. The prediction is
\begin{equation}
 \hat{y}(x)=T(x)+\beta\,\exp\!\left[-\lambda\max\!\left(0,d(x)-d_{q}\right)\right]r(x).
 \label{eq:model}
\end{equation}
Here $d(x)$ measures how far catalyst $x$ lies from the centre of the fitting catalysts in feature
space: it is the root-mean-square
of the features of $x$ after each feature has been standardized with the mean and standard deviation
of the fitting catalysts. The value $d_q$ is the $q$ quantile of $d$ over the fitting catalysts, $\beta$
scales the residual term and $\lambda$ sets how strongly the residual term is reduced for catalysts
beyond $d_q$ \cite{Schultz2025AD}. In setting~1, the residual learner of PR-ML (Ridge) is ridge regression \cite{Ridge1970} on
standardized features with a penalty weight $\alpha=0.1$ on the squared coefficients, the value selected
for Ridge used alone, and $\beta=0.5$, $\lambda=1$ and $q=0.95$. In setting~2 and in FastCat and DASH, and for every GP residual learner,
$\beta=1$ and $\lambda=0$, so the residual term is added in full. Used on its own, the Tafel term is Eq.~\ref{eq:model} with $\beta=0$ and uses no endpoint labels. \mbox{PR-ML}
with a given residual learner is denoted PR-ML (learner), for example PR-ML (Ridge), with a setting
added after a comma where needed, for example PR-ML (Ridge, $\alpha=10$); a learner used alone to
predict the endpoint is denoted by its name, for example Ridge.

In setting~2, the residual learner of PR-ML (ExtraTrees) is extremely randomized trees
\cite{ExtraTrees2006} with 64 trees; each tree is fitted on all labelled catalysts, every split
considers all features and a leaf may contain a single catalyst. In FastCat and DASH, four residual
learners were used: ExtraTrees with the same settings, a GP, and ridge regression ($\alpha=0.1$) on
composition and measured values or on the measured values only, which give PR-ML (Ridge) and PR-ML
(Ridge, measured values only).

The GPs \cite{Rasmussen2006GP} use a Mat\'ern covariance ($\nu=5/2$) with one length scale per feature,
multiplied by a variance factor, plus a white-noise term. Features are standardized
with the fitting catalysts, and the variance, length scales and noise level are chosen by maximizing
the marginal likelihood, the probability of the labels under the GP, with two restarts of the optimizer.
With a GP as the residual learner, two versions were used. In PR-ML (GP, mean-residual prior mean), the
labelled residuals are centred on their mean and scaled to unit variance before fitting, so that the prior mean is the mean of $y-T$ over the labelled catalysts (the mean labelled
residual) and, far from the labelled catalysts,
the prediction returns to the Tafel term plus this value; this version is used with labels from the new
library and in FastCat and DASH. In PR-ML (GP, zero prior mean), the prior mean is zero, so the
prediction returns
to the Tafel term; this version is used in setting~1 (Table~\ref{tab:setting1}) and is also reported
for setting~2 (Supplementary Table~\ref{tab:si_twelve}). The GP also gives a predictive standard deviation, from
which 90\% prediction intervals (mean $\pm1.645$ standard deviations) are obtained.

\subsection*{Compared methods}

The compared methods represent the approaches A, B and C of the Introduction that apply to these data.
The data-driven models learn the endpoint directly from composition, measured values or both. Ridge represents a
linear relation. ExtraTrees and histogram gradient boosting (HGB) are tree
ensembles, whose predictions are averages of training labels (ExtraTrees) or sums of fitted tree
outputs (HGB). A multilayer perceptron (MLP) represents a flexible nonlinear relation. These models were
implemented in scikit-learn \cite{Pedregosa2011}. GP (composition) is of the type used in the original
Ni--Pd--Pt--Ru study \cite{Thelen2025OER}, and GP (composition and measured currents) uses composition
and measured values, the same inputs as PR-ML.

Extrapolations without a catalyst-dependent correction test the Tafel equation alone. They are the
Tafel term; the Tafel term with a constant offset, that is, the Tafel term plus the mean labelled
residual, which in setting~2 equals the 1.55~V$_\mathrm{RHE}$ current multiplied by one factor fitted on
the labelled catalysts (Fig.~\ref{fig:setting2}a); the Tafel term (median slope of the training
catalysts), which uses only $g$ ($w=0$); the Tafel equation with a standard slope of 40, 60 or
120~mV~dec$^{-1}$ (standard Tafel); and the
1.55~V$_\mathrm{RHE}$ current as the prediction (Fig.~\ref{fig:slope}). The fixed slope
(240~mV~dec$^{-1}$, chosen after the endpoints had been seen) is shown for reference. In the
twelve-library comparison of Fig.~\ref{fig:slope}b, the slope of the Tafel term (library slope) is the
median of the two-point slopes between 1.40 and 1.55~V$_\mathrm{RHE}$ of all catalysts of the predicted
library. In FastCat and DASH, standard Tafel (60, 80 and 120~mV~dec$^{-1}$) was also computed; for
each dataset, the slope with the lowest error on the evaluated catalysts (chosen after the endpoints had
been seen) is reported as standard Tafel matched to that dataset.

The Tafel term with affine calibration and the Tafel term with isotonic calibration test whether one
function applied to the Tafel term of every catalyst can replace the residual term. The affine calibration replaces the Tafel term $T$ by $a+cT$ with $c\geq 0$. The isotonic
calibration replaces $T$ by a non-decreasing function of $T$ fitted by isotonic regression, extended by
straight lines fitted to the lowest and highest 10\% of the Tafel-term predictions used for fitting; this fraction was
chosen among 0.1, 0.2, 0.3 and 0.5 by leave-one-library-out error on libraries 3--5. Both calibrations
were fitted, for two and for three measured currents, to the Tafel-term predictions and measured endpoints
of libraries 3, 4 and 5, each predicted from the libraries before it, and then applied to the Ni--Ru
predictions.

In every comparison, all models used the same labelled catalysts and the same measured values of the
predicted catalysts. The learned models used the same features as PR-ML, except GP (composition) and Ridge (measured values
only). In setting~1, the slope of the Tafel term
also uses the currents of the training catalysts at 17 potentials between 1.40 and
1.55~V$_\mathrm{RHE}$; the compared models do not use these currents. In the paired comparisons of FastCat and DASH, PR-ML and the learner used alone had the same features, hyperparameters and labelled
catalysts.

\subsection*{Selection of hyperparameters}

In setting~1, the hyperparameters of PR-ML, with Ridge as the residual learner, and of each
data-driven model were selected only on
libraries 1--5. Each of these libraries was predicted from the other four with 2, 3, 5, 9 and 17
measured currents at evenly spaced potentials between 1.40 and 1.55~V$_\mathrm{RHE}$. For each
predicted library and number of measured currents, the configurations of a model were ranked by four
criteria. The first two were the MAE and the Spearman correlation \cite{Spearman1904} between predicted
and measured endpoints. The third was the fraction of the 20\% most active catalysts that were among the
20\% predicted to be most active. The fourth was the difference between the mean measured activity of
these two groups of catalysts. The configuration with the best mean rank was recorded,
which gave 25 recorded configurations per model. Their values were averaged and rounded to the nearest
value of the search grid. Penalty weights were averaged on a logarithmic scale. For PR-ML, $\alpha$ and
$\lambda$ were averaged only over the recorded configurations that contained a residual term
($\beta>0$), and $q$ only over those with $\lambda>0$. The hidden-layer sizes of the MLP were set to the
most frequent choice. For PR-ML, $w$ and the number of highest measured potentials used for $s_\mathrm{c}$ were selected first,
and $\alpha$, $\beta$, $\lambda$ and $q$ second. The search grids were $w\in\{0,0.1,0.25,0.5,0.75,1\}$,
the two or three highest measured potentials or all measured potentials for $s_\mathrm{c}$, $\alpha\in\{1,10,100\}$,
$\beta\in\{0,0.25,0.5,0.75,1\}$, $\lambda\in\{0,0.25,1,2,5,10\}$ and $q\in\{0.90,0.95,0.99\}$ for PR-ML; $\alpha\in\{0.01,0.1,1,10,100\}$ for Ridge; a minimum leaf size of 1, 2, 4, 8 or 16 and
a feature fraction of 0.5, 0.8 or 1.0 for ExtraTrees; a penalty of 0.1, 1 or 10 and a minimum leaf size
of 10, 20 or 40 for HGB; and hidden layers of (16), (32, 16) or (32, 32) units with a weight penalty of
0.001, 0.01 or 0.1 for the MLP. The numbers of trees and boosting iterations were fixed at 300.

For PR-ML, this procedure gave $w=0.25$, two potentials for $s_\mathrm{c}$, $\alpha=10$,
$\beta=0.5$, $\lambda=1$ and $q=0.95$. For the data-driven models, it gave Ridge with $\alpha=0.1$; ExtraTrees with at least 8 training catalysts per leaf and a random 80\% of
the features considered at each split; HGB with a penalty of 1 on the squared leaf values and at least 20 catalysts per leaf; and an MLP with two hidden layers of 32
units, a weight penalty of 0.01 and early stopping. No catalyst of library~6 or of libraries 7--12
entered these choices. For the comparisons, PR-ML (Ridge) used the penalty weight $\alpha=0.1$ selected for
Ridge, so that the two models had the same learner, penalty weight, features and
training catalysts; this penalty weight was also selected on libraries 1--5 only. PR-ML (Ridge,
$\alpha=10$), which uses the penalty weight selected for PR-ML itself, had an MAE of 0.203 instead of
0.194~\si{\milliamperepercmsquared} for all Ni--Ru catalysts (Supplementary Table~\ref{tab:si_budget}). In setting~1, the predictions of ExtraTrees, HGB and the MLP were averaged over
five runs with different random seeds, and the GPs were fitted once.

In setting~2 and in FastCat and DASH, all hyperparameters were fixed before the analysis and were the
same for every library, dataset, design and number of labels; ExtraTrees had the same settings as the
residual learner of PR-ML (ExtraTrees). The GP kernel parameters were fitted to the labelled
catalysts of each draw.

\subsection*{Setting 1: fitting on other libraries}

PR-ML and the compared methods were fitted on libraries 1--5 and evaluated on library~6 (Ni--Ru),
and fitted on libraries 1--6 with the same hyperparameters and evaluated on libraries 7--12. Catalysts
were grouped by $z=(\ell-\ell_\mathrm{max})/\mathrm{IQR}$, where $\ell$ is $\log_{10}$ of the endpoint
current density, and $\ell_\mathrm{max}$ and IQR are the maximum and interquartile range of $\ell$ over
the training catalysts. Catalysts with $z>0$ lie above the training range, and those with $z>2$ lie far
above it (144 of the 322 Ni--Ru catalysts).

For the selection of catalysts above the training range, a catalyst was selected when its current at
1.55~V$_\mathrm{RHE}$ exceeded the highest 1.55~V$_\mathrm{RHE}$ current of the training libraries. The
rule was applied with libraries 1--5 as training libraries and, for each library, with the other eleven
libraries (Supplementary Table~\ref{tab:si_selection}); in the latter case, the hyperparameters selected
on libraries 1--5 were kept. For the selected catalysts, PR-ML (Ridge) was compared with the Tafel term, Ridge
and ExtraTrees with two, three and five measured currents; Ridge was the more accurate of the two
data-driven models on these catalysts.

For the paired comparisons of setting~1, PR-ML (Ridge) and PR-ML (ExtraTrees) used Ridge and
ExtraTrees as residual learners with the hyperparameters selected for these models used alone
($\alpha=0.1$ for Ridge) and with $\beta$, $\lambda$ and $q$ of setting~1, and each was compared with the same learner used alone. These models were evaluated with 1, 2, 3, 5, 9 and 17 measured
currents; with one measured current, the measured potential was 1.475~V$_\mathrm{RHE}$. Four further
analyses used these models with the same hyperparameters. In the first, $g$ was estimated only from the
currents of the training catalysts at the measured potentials. In the second, the models were fitted on
35, 50, 85, 169, 337 or 674 of the 1,685 catalysts of libraries 1--5, chosen without regard to their
endpoints. For each of 20 random subsets, the catalysts of each library were put in a random order
and the required number was taken from the start of each library, so that every library contributed the
same fraction and each smaller set was contained in each larger one. In the third, the positions of the
measured potentials were varied: two, three, five or nine measured potentials were evenly spaced, taken
as consecutive potentials at the low or at the high end of the 17 potentials, or chosen at random (12
random sets for each number, the same for all models). In the fourth, normally distributed errors with
a standard deviation of 0.01, 0.03, 0.05 or 0.10 were added to the logarithms of the measured currents
of the evaluated catalysts in 12 repetitions for each standard deviation, and every model received the same
perturbed values; with nine and seventeen measured currents, $s_\mathrm{c}$ was also obtained by a
least-squares fit over the three highest or all measured potentials.

\subsection*{Setting 2: fitting on a few endpoints of the new library}

In setting~2, each library in turn was the new library, and only its own catalysts were used. PR-ML
was fitted on the labels of 3, 5, 10, 20 or 40 catalysts, drawn either at random from
the whole library or only from the half of the library with the lower endpoint currents. The second
choice is called range-shifted selection, because the labels then cover only the lower part of the
activity range. For each number of labels, 50 draws were made, all models used the same
labelled catalysts in each draw, and every unlabelled catalyst of the library was evaluated. PR-ML
(ExtraTrees) was compared with ExtraTrees, Ridge, GP (composition), GP (composition and
measured currents) and the Tafel term with a constant offset; Ridge used $\alpha=0.1$, as in
setting~1. PR-ML (GP, mean-residual prior mean) and
PR-ML (GP, zero prior mean) were fitted to the same draws.

For the comparison with the most accurate of three compared methods
(Supplementary Table~\ref{tab:si_stop155}), the measured potentials were 1.40 and 1.55, or 1.50 and
1.55~V$_\mathrm{RHE}$, and the endpoint was 1.70~V$_\mathrm{RHE}$. Catalysts were divided into those
below and those above the largest label of each draw. For each combination of measured potentials,
selection, number of labels and group of catalysts, the compared method was the one of the Tafel
term (library slope), Ridge and ExtraTrees with the lowest mean error over the twelve libraries on the
evaluated catalysts. The residual learner of PR-ML was whichever of Ridge and ExtraTrees had the
lower mean error when used alone.

\subsection*{Tests above the label range in FastCat and DASH}

The catalysts of each dataset were sorted by their measured endpoint. The labelled catalysts were drawn
from the less active 50\% or 70\% of the catalysts (the label fraction), and the remaining, more active
catalysts were evaluated. For each dataset, design, label fraction and number of labelled catalysts (3,
5, 10, 20 or 40), the labelled catalysts were drawn at random 200 times, and all models used the same
labelled catalysts in each draw. ExtraTrees, GP, Ridge and Ridge (measured values only) were also used alone
to predict the endpoint, and the Tafel term with a constant offset was also compared. The GP, as
residual learner and used alone, was applied to the two designs
reported in the main text, and the other learners to all 16 combinations of design and endpoint
(Supplementary Table~\ref{tab:si_combinations}).

In a second test, the catalysts were divided by the order in which they were made. In FastCat, whose
catalysts were made and measured in a sequence of experiments, the labelled catalysts were drawn from
the 60 earliest of the 89 catalysts and the 29 later catalysts were evaluated. In DASH, they were drawn
from synthesis batches 1 and 2 (51 catalysts), and the catalysts of batches 3--5 (84) were evaluated.
The numbers of labelled catalysts, the 200 draws and the models were those of the test above the label
range (Supplementary Table~\ref{tab:si_chrono}). An additional DASH test, in which the current at a higher potential is predicted and the catalysts
are divided by activity within each synthesis batch, is described with its own model choices in
\hyperref[note:si_dash]{Supplementary Note~1} and Supplementary Table~\ref{tab:si_dash}.

\subsection*{Evaluation metrics and statistical analysis}

The models predict $\log_{10} j$ or the overpotential. The MAE is computed in
\si{\milliamperepercmsquared} after converting the predictions to current density in setting~1
(Table~\ref{tab:setting1}) and for the Ni--Ru library in setting~2, in $\log_{10}$ current for the
twelve-library comparisons (Fig.~\ref{fig:slope}b and setting~2) and in mV for FastCat and DASH. For
the twelve libraries, the MAE was computed for each library, averaged over the draws in setting~2, and
then averaged over the libraries. Because a new library may lie inside or above the training range, the larger of the two
setting-1 errors (library~6 and libraries 7--12) summarizes each model in Table~\ref{tab:setting1}. The
Spearman correlation between predicted and measured endpoints shows whether the catalysts are ranked
correctly, and the median ratio of predicted to measured current shows whether the predictions are
systematically too low or too high. For each draw, we also computed the fraction of the unlabelled
catalysts more active than the best labelled catalyst that each model predicts to be more active than
the best labelled catalyst. This fraction is zero for a draw without such catalysts and was averaged
over draws and, where stated, over libraries.

Differences in library MAE between two models in setting~2 were tested with the two-sided Wilcoxon
signed-rank test \cite{Wilcoxon1945} over the twelve libraries. In FastCat and DASH, the absolute error
of each evaluated catalyst was averaged over the draws. Two kinds of pairs were compared: PR-ML
against the same learner used alone, and PR-ML or the Tafel term with a constant offset against the
Tafel term (dataset median slope). For each pair, the MAE of the first model minus that of the second
was computed on 2,000 bootstrap resamples of the evaluated catalysts, drawn with replacement and the
same for both models \cite{Efron1979}. The 95\% confidence interval runs from the 2.5th to the 97.5th
percentile of these differences. The first model had a lower error when the whole interval was below
zero, a higher error when it was above zero, and no significant
difference otherwise. The confidence intervals of single MAEs (Fig.~\ref{fig:independent}a) were
obtained from the same resamples. For Supplementary Table~\ref{tab:si_stop155}, the interval was
obtained from 10,000 bootstrap resamples of the twelve library-level differences. In setting~1, the 95\%
confidence intervals of paired differences were obtained in the same way from 2,000 bootstrap resamples
of the evaluated catalysts. The evaluated catalysts were also resampled by composition: they were
grouped into 8, 12 or 16 clusters by $k$-means clustering of their Ni, Pd, Pt and Ru fractions
\cite{Lloyd1982}, and whole clusters were drawn with replacement 5,000 times, with the measured endpoint
and both predictions of each catalyst kept together. For the reduced training sets, PR-ML was taken to have a lower (higher) error when the 2.5th and 97.5th percentiles of the paired differences over
the 20 random subsets were both below (above) zero. For the added errors, the median difference over the
12 repetitions and the number of repetitions in which PR-ML had the smaller MAE are reported. No adjustment for
multiple comparisons was applied, and the counts are reported over all conditions.

To check the implementation, a separately written program repeated the FastCat and DASH tests for the
two designs of the main text, a label fraction of 50\% and 5 and 10 labelled catalysts, with 100 new
random draws and only the data loading shared. It gave identical errors for the Tafel term, and
the same outcome (lower error, no significant difference or higher error) in 22 of the 24 paired
comparisons of PR-ML (Ridge), PR-ML (Ridge, measured values only) and PR-ML (ExtraTrees) with the same
learner used alone. All computations ran on CPUs with Python 3.13, NumPy 2.4,
pandas 2.3, SciPy 1.17 and scikit-learn 1.8, and the random seeds are recorded with the results.

\section*{Data availability}

The Ni--Pd--Pt--Ru data are available under CC BY 4.0 at Zenodo
(\url{https://doi.org/10.5281/zenodo.14891704}), FastCat version 1 under the MIT licence
(\url{https://doi.org/10.11583/DTU.28494185.v1}) and DASH under CC0
(\url{https://doi.org/10.5061/dryad.nk98sf83g}). The processed tables, the predictions for each
catalyst and the source data of the figures will be deposited in a public repository, and the DOI will
be provided before publication.

\section*{Code availability}

The code is available from the corresponding author on reasonable request.

\clearpage
\bibliographystyle{unsrtnat}
\bibliography{references}

\clearpage
\setcounter{page}{1}
\renewcommand{\thepage}{S\arabic{page}}
\setcounter{table}{0}
\renewcommand{\tablename}{Supplementary Table}
\renewcommand{\thetable}{\arabic{table}}
\renewcommand{\theHtable}{S\arabic{table}}
\renewcommand{\floatpagefraction}{0.8}
\renewcommand{\topfraction}{0.9}
\renewcommand{\bottomfraction}{0.6}
\renewcommand{\textfraction}{0.07}
\makeatletter\setlength{\@fptop}{0pt}\makeatother

\begin{center}
{\Large\bfseries Supplementary Information}
\end{center}
\phantomsection\label{sec:si}

\section*{Supplementary Tables}

PR-ML denotes the physics-residual machine-learning model of the main text: the Tafel term plus a
residual term learned by the residual learner named in parentheses. In Supplementary
Tables~\ref{tab:si_cluster}--\ref{tab:si_reduction}, PR-ML (Ridge) is the model of
Table~\ref{tab:setting1}, and PR-ML (ExtraTrees) uses ExtraTrees as the residual learner. Both use the
hyperparameters of the learner used alone, with the residual term halved and reduced for catalysts far
from the training data ($\beta=0.5$, $\lambda=1$, $q=0.95$). PR-ML (Ridge, $\alpha=10$) in
Supplementary Table~\ref{tab:si_budget} uses the penalty weight selected for PR-ML itself
(\hyperref[sec:methods]{Methods}). Far above: the 144 Ni--Ru catalysts whose logarithmic current at
1.70~V$_\mathrm{RHE}$ exceeds the training maximum by more than two interquartile ranges of the
logarithmic training currents. All values in these tables refer to setting~1, with models fitted on
libraries 1--5 unless stated.

\begin{table}[!htbp]
\centering
\caption{\textbf{Paired differences in setting~1 with resampling by composition.} Difference in MAE
(\si{\milliamperepercmsquared}) between the first and the second model of each comparison (negative: the first model is more
accurate), with two and three measured currents. The 95\% confidence intervals come from 5,000
resamples of whole clusters of catalysts; the clusters were obtained by $k$-means clustering of the Ni,
Pd, Pt and Ru fractions into 8, 12 or 16 clusters. For the catalysts far above the training range, the
interval lay below zero in all 18 cases (1, 2, 3, 5, 9 and 17 measured currents and three numbers of
clusters) for both comparisons with a learner used alone. For all Ni--Ru catalysts, it lay below zero
in 14 and 17 of the 18 cases for these two comparisons, and there was no significant difference
in the other cases. Against the Tafel term, the interval lay below zero for both models and both
groups of catalysts with two and three measured currents.}
\label{tab:si_cluster}
\small
\setlength{\tabcolsep}{5pt}
\begin{tabular}{lcccc}
\toprule
 & & \multicolumn{3}{c}{95\% confidence interval} \\
\cmidrule(lr){3-5}
Number of measured currents & Difference & 8 clusters & 12 clusters & 16 clusters \\
\midrule
\multicolumn{5}{l}{\textbf{Far above (144)}} \\
\multicolumn{5}{l}{\quad PR-ML (Ridge) vs Ridge} \\
\qquad 2 & $-2.309$ & $-2.60$ to $-1.93$ & $-2.55$ to $-2.00$ & $-2.52$ to $-2.03$ \\
\qquad 3 & $-1.452$ & $-1.92$ to $-0.82$ & $-1.85$ to $-0.90$ & $-1.79$ to $-0.98$ \\
\multicolumn{5}{l}{\quad PR-ML (ExtraTrees) vs ExtraTrees} \\
\qquad 2 & $-2.799$ & $-3.14$ to $-2.36$ & $-3.08$ to $-2.44$ & $-3.04$ to $-2.47$ \\
\qquad 3 & $-2.806$ & $-3.22$ to $-2.27$ & $-3.16$ to $-2.34$ & $-3.10$ to $-2.40$ \\
\multicolumn{5}{l}{\quad PR-ML (Ridge) vs Tafel term} \\
\qquad 2 & $-0.559$ & $-0.64$ to $-0.48$ & $-0.63$ to $-0.49$ & $-0.62$ to $-0.50$ \\
\qquad 3 & $-0.351$ & $-0.38$ to $-0.32$ & $-0.38$ to $-0.32$ & $-0.37$ to $-0.33$ \\
\multicolumn{5}{l}{\quad PR-ML (ExtraTrees) vs Tafel term} \\
\qquad 2 & $-0.171$ & $-0.18$ to $-0.16$ & $-0.18$ to $-0.16$ & $-0.18$ to $-0.16$ \\
\qquad 3 & $-0.087$ & $-0.09$ to $-0.08$ & $-0.09$ to $-0.08$ & $-0.09$ to $-0.08$ \\
\midrule
\multicolumn{5}{l}{\textbf{All Ni--Ru (322)}} \\
\multicolumn{5}{l}{\quad PR-ML (Ridge) vs Ridge} \\
\qquad 2 & $-1.136$ & $-1.88$ to $-0.20$ & $-1.75$ to $-0.44$ & $-1.68$ to $-0.46$ \\
\qquad 3 & $-0.614$ & $-1.26$ to $+0.16$ & $-1.18$ to $-0.01$ & $-1.15$ to $-0.01$ \\
\multicolumn{5}{l}{\quad PR-ML (ExtraTrees) vs ExtraTrees} \\
\qquad 2 & $-1.466$ & $-2.33$ to $-0.40$ & $-2.17$ to $-0.68$ & $-2.08$ to $-0.69$ \\
\qquad 3 & $-1.431$ & $-2.32$ to $-0.31$ & $-2.18$ to $-0.63$ & $-2.12$ to $-0.64$ \\
\multicolumn{5}{l}{\quad PR-ML (Ridge) vs Tafel term} \\
\qquad 2 & $-0.347$ & $-0.49$ to $-0.15$ & $-0.47$ to $-0.20$ & $-0.45$ to $-0.20$ \\
\qquad 3 & $-0.207$ & $-0.30$ to $-0.08$ & $-0.29$ to $-0.11$ & $-0.28$ to $-0.12$ \\
\multicolumn{5}{l}{\quad PR-ML (ExtraTrees) vs Tafel term} \\
\qquad 2 & $-0.125$ & $-0.16$ to $-0.08$ & $-0.16$ to $-0.09$ & $-0.15$ to $-0.09$ \\
\qquad 3 & $-0.079$ & $-0.10$ to $-0.06$ & $-0.10$ to $-0.06$ & $-0.09$ to $-0.06$ \\
\bottomrule
\end{tabular}
\end{table}

\begin{table}[!htbp]
\centering
\caption{\textbf{Slope of the Tafel term with and without the currents at 17 potentials.} MAE
(\si{\milliamperepercmsquared}) with the median slope $g$ estimated from the currents of the training catalysts at 17
potentials between 1.40 and 1.55~V$_\mathrm{RHE}$, as in the main analysis, or only from their currents
at the measured potentials. With 17 measured currents the two estimates are identical (values in
Supplementary Table~\ref{tab:si_budget}). With the slope from the measured potentials only, PR-ML (Ridge)
and PR-ML (ExtraTrees) had a lower error than the same learner used alone (95\% confidence intervals
from 2,000 bootstrap resamples of the catalysts) in all 15 combinations of the number of measured
currents (2, 3, 5, 9 and 17) and the group of catalysts (far above, above the training range and all); the group above the training range (282 catalysts) is not shown. The slope from 17 potentials
gave the lower MAE with two, three and five measured currents (for example 0.177 instead of 0.709 for
PR-ML (Ridge) far above the training range), significantly in every case with two and three and in 8 of 9
cases with five, and, with nine measured currents, a higher error for PR-ML (Ridge).}
\label{tab:si_slope}
\small
\setlength{\tabcolsep}{4.5pt}
\begin{tabular}{lcccccccc}
\toprule
 & \multicolumn{4}{c}{Far above (144)} & \multicolumn{4}{c}{All Ni--Ru (322)} \\
\cmidrule(lr){2-5}\cmidrule(lr){6-9}
Number of measured currents & 2 & 3 & 5 & 9 & 2 & 3 & 5 & 9 \\
\midrule
\multicolumn{9}{l}{Tafel term} \\
\quad slope from 17 potentials & 0.735 & 0.620 & 0.261 & 0.179 & 0.541 & 0.536 & 0.311 & 0.231 \\
\quad slope from the measured potentials & 1.561 & 1.426 & 0.644 & 0.235 & 1.011 & 1.021 & 0.539 & 0.269 \\
\multicolumn{9}{l}{PR-ML (Ridge)} \\
\quad slope from 17 potentials & 0.177 & 0.269 & 0.237 & 0.257 & 0.194 & 0.329 & 0.230 & 0.196 \\
\quad slope from the measured potentials & 0.709 & 0.912 & 0.266 & 0.209 & 0.477 & 0.714 & 0.290 & 0.184 \\
Ridge & 2.486 & 1.721 & 2.157 & 2.216 & 1.330 & 0.943 & 1.098 & 1.118 \\
\multicolumn{9}{l}{PR-ML (ExtraTrees)} \\
\quad slope from 17 potentials & 0.565 & 0.532 & 0.216 & 0.176 & 0.416 & 0.457 & 0.259 & 0.204 \\
\quad slope from the measured potentials & 1.238 & 1.216 & 0.539 & 0.190 & 0.781 & 0.849 & 0.445 & 0.219 \\
ExtraTrees & 3.363 & 3.338 & 3.318 & 3.307 & 1.882 & 1.888 & 1.875 & 1.871 \\
\bottomrule
\end{tabular}
\end{table}

\begin{table}[!htbp]
\centering
\caption{\textbf{Positions of the measured potentials, errors added to the measured currents, and
potentials used for the catalyst's own slope} (catalysts far above the training range; MAE differences
in \si{\milliamperepercmsquared}). \textbf{a}, Positions: two, three, five or nine measured potentials evenly spaced, taken
as consecutive potentials at the low or at the high end of the 17 potentials, or chosen at random (12
random sets for each number), which gives 60 sets of measured potentials. Number of sets in which the
first model had a lower error / no significant difference / a higher error (95\% confidence intervals
from 2,000 bootstrap resamples of the catalysts), and the smallest reduction in MAE relative to the
learner used alone over the 60 sets (--, not applicable). With the measured potentials at the high end
of the window, the Tafel term gave an MAE of 0.172~\si{\milliamperepercmsquared} for two, three, five and nine
measured potentials; PR-ML (ExtraTrees) showed no significant difference from it, and PR-ML (Ridge) had
a higher error. \textbf{b}, Errors: normally distributed
errors with the given standard deviation (SD, in $\log_{10}$ current) added to the measured currents
of the evaluated catalysts. Median over 12 repetitions of the difference in MAE between PR-ML and
the same learner used alone, with the number of repetitions in which PR-ML had the smaller MAE in
parentheses; SD 0, without added errors (one evaluation). The catalyst's own slope $s_\mathrm{c}$ is
taken between the two highest measured potentials, as in the main analysis. \textbf{c}, Potentials
used for $s_\mathrm{c}$ with nine and seventeen measured currents, whose spacing is 19 and 9~mV:
two highest, three highest or all measured potentials, with an independent set of 12 repetitions for
each SD; values are median differences over the 12 repetitions with the number of repetitions in which
PR-ML had the smaller MAE in parentheses, as in \textbf{b}. With $s_\mathrm{c}$ taken over all
measured potentials, PR-ML had the smaller MAE in all 12 repetitions at every SD.}
\label{tab:si_robust}
\small
\textbf{a}\par\smallskip
\setlength{\tabcolsep}{4pt}
\begin{tabular}{lcc}
\toprule
Comparison & \makecell{Lower / no significant\\difference / higher} & \makecell{Smallest\\reduction} \\
\midrule
PR-ML (Ridge) vs Ridge & 60 / 0 / 0 & 0.545 \\
PR-ML (ExtraTrees) vs ExtraTrees & 60 / 0 / 0 & 0.991 \\
PR-ML (Ridge) vs Tafel term & 16 / 5 / 39 & -- \\
PR-ML (ExtraTrees) vs Tafel term & 26 / 20 / 14 & -- \\
\bottomrule
\end{tabular}

\bigskip
\textbf{b}\par\smallskip
\setlength{\tabcolsep}{4pt}
\begin{tabular}{lccccc}
\toprule
Number of measured currents & SD 0 & SD 0.01 & SD 0.03 & SD 0.05 & SD 0.10 \\
\midrule
\multicolumn{6}{l}{PR-ML (Ridge) vs Ridge} \\
\quad 2 & $-2.309$ & $-2.270$ (12) & $-2.088$ (12) & $-1.888$ (12) & $-1.338$ (12) \\
\quad 3 & $-1.452$ & $-1.382$ (12) & $-1.194$ (12) & $-0.928$ (12) & $-0.207$ (10) \\
\quad 5 & $-1.920$ & $-1.837$ (12) & $-1.438$ (12) & $-1.054$ (12) & $-0.699$ (12) \\
\quad 9 & $-1.959$ & $-1.832$ (12) & $-1.180$ (12) & $-0.441$ (11) & $+0.776$ (0) \\
\quad 17 & $-1.983$ & $-1.617$ (12) & $-0.311$ (11) & $+1.232$ (0) & $+6.947$ (0) \\
\multicolumn{6}{l}{PR-ML (ExtraTrees) vs ExtraTrees} \\
\quad 2 & $-2.799$ & $-2.797$ (12) & $-2.706$ (12) & $-2.545$ (12) & $-1.993$ (12) \\
\quad 3 & $-2.806$ & $-2.792$ (12) & $-2.663$ (12) & $-2.440$ (12) & $-1.612$ (12) \\
\quad 5 & $-3.102$ & $-2.993$ (12) & $-2.645$ (12) & $-2.206$ (12) & $-1.207$ (12) \\
\quad 9 & $-3.131$ & $-2.941$ (12) & $-2.289$ (12) & $-1.555$ (12) & $+0.707$ (0) \\
\quad 17 & $-3.121$ & $-2.675$ (12) & $-1.367$ (12) & $+0.258$ (3) & $+7.226$ (0) \\
\bottomrule
\end{tabular}

\bigskip
\textbf{c}\par\smallskip
\begin{tabular}{llccccc}
\toprule
Number of measured currents & Potentials for $s_\mathrm{c}$ & SD 0 & SD 0.01 & SD 0.03 & SD 0.05 & SD 0.10 \\
\midrule
\multicolumn{7}{l}{PR-ML (Ridge) vs Ridge} \\
\quad 9 & two highest & $-1.959$ & $-1.813$ (12) & $-1.199$ (12) & $-0.498$ (12) & $+0.791$ (0) \\
\quad 9 & three highest & $-2.002$ & $-1.909$ (12) & $-1.473$ (12) & $-1.082$ (12) & $-0.793$ (12) \\
\quad 9 & all 9 & $-1.769$ & $-1.753$ (12) & $-1.563$ (12) & $-1.381$ (12) & $-1.731$ (12) \\
\quad 17 & two highest & $-1.983$ & $-1.601$ (12) & $-0.387$ (12) & $+1.141$ (0) & $+9.891$ (0) \\
\quad 17 & three highest & $-2.003$ & $-1.850$ (12) & $-1.194$ (12) & $-0.511$ (12) & $+0.648$ (0) \\
\quad 17 & all 17 & $-1.729$ & $-1.723$ (12) & $-1.558$ (12) & $-1.383$ (12) & $-1.764$ (12) \\
\multicolumn{7}{l}{PR-ML (ExtraTrees) vs ExtraTrees} \\
\quad 9 & two highest & $-3.131$ & $-2.925$ (12) & $-2.331$ (12) & $-1.545$ (12) & $+0.777$ (1) \\
\quad 9 & three highest & $-3.089$ & $-2.986$ (12) & $-2.620$ (12) & $-2.180$ (12) & $-1.146$ (12) \\
\quad 9 & all 9 & $-2.607$ & $-2.605$ (12) & $-2.567$ (12) & $-2.411$ (12) & $-1.972$ (12) \\
\quad 17 & two highest & $-3.121$ & $-2.657$ (12) & $-1.434$ (12) & $+0.195$ (4) & $+10.509$ (0) \\
\quad 17 & three highest & $-3.123$ & $-2.917$ (12) & $-2.295$ (12) & $-1.554$ (12) & $+0.574$ (1) \\
\quad 17 & all 17 & $-2.564$ & $-2.569$ (12) & $-2.520$ (12) & $-2.394$ (12) & $-2.033$ (12) \\
\bottomrule
\end{tabular}
\end{table}

\begin{table}[!htbp]
\centering
\caption{\textbf{Number of measured currents in setting~1.} MAE (\si{\milliamperepercmsquared}) with 1, 2, 3, 5, 9 and 17
measured currents per catalyst between 1.40 and 1.55~V$_\mathrm{RHE}$. The single measured current is
at 1.475~V$_\mathrm{RHE}$, so its distance to the endpoint (0.225~V) differs from that with two or more
measured currents (0.15~V). Ni--Ru: fitted on libraries 1--5; libraries 7--12: fitted on libraries
1--6. Larger of two: the larger of the MAEs over all Ni--Ru catalysts and over libraries 7--12. With
95\% confidence intervals from 2,000 bootstrap resamples of the catalysts, the three PR-ML models
had a lower error than the same learner used alone with every number of measured currents in the Ni--Ru
library, for both groups of catalysts. For the catalysts far above the training range, they had a lower
error than the Tafel term with two and three measured currents; with nine and seventeen, the Tafel
term was more accurate than PR-ML (Ridge) and PR-ML (Ridge, $\alpha=10$), with no significant difference
from PR-ML (ExtraTrees). In libraries 7--12, PR-ML (ExtraTrees) had a higher error than ExtraTrees with
every number of measured currents, and PR-ML (Ridge) had a lower error
than Ridge with 2 and 5 measured currents, a higher error with
1 and 17, and no significant difference with 3 and 9.}
\label{tab:si_budget}
\small
\setlength{\tabcolsep}{5pt}
\begin{tabular}{lcccccc}
\toprule
 & \multicolumn{6}{c}{Number of measured currents} \\
\cmidrule(lr){2-7}
 & 1 & 2 & 3 & 5 & 9 & 17 \\
\midrule
\multicolumn{7}{l}{\textbf{Far above (144)}} \\
\quad Tafel term & 2.275 & 0.735 & 0.620 & 0.261 & 0.179 & 0.172 \\
\quad PR-ML (Ridge) & 2.434 & 0.177 & 0.269 & 0.237 & 0.257 & 0.257 \\
\quad Ridge & 2.969 & 2.486 & 1.721 & 2.157 & 2.216 & 2.240 \\
\quad PR-ML (ExtraTrees) & 2.284 & 0.565 & 0.532 & 0.216 & 0.176 & 0.178 \\
\quad ExtraTrees & 3.381 & 3.363 & 3.338 & 3.318 & 3.307 & 3.298 \\
\quad PR-ML (Ridge, $\alpha=10$) & 2.429 & 0.191 & 0.266 & 0.234 & 0.253 & 0.254 \\
\midrule
\multicolumn{7}{l}{\textbf{All Ni--Ru (322)}} \\
\quad Tafel term & 1.504 & 0.541 & 0.536 & 0.311 & 0.231 & 0.210 \\
\quad PR-ML (Ridge) & 1.519 & 0.194 & 0.329 & 0.230 & 0.196 & 0.187 \\
\quad Ridge & 1.693 & 1.330 & 0.943 & 1.098 & 1.118 & 1.131 \\
\quad PR-ML (ExtraTrees) & 1.478 & 0.416 & 0.457 & 0.259 & 0.204 & 0.190 \\
\quad ExtraTrees & 1.933 & 1.882 & 1.888 & 1.875 & 1.871 & 1.870 \\
\quad PR-ML (Ridge, $\alpha=10$) & 1.516 & 0.203 & 0.321 & 0.245 & 0.209 & 0.193 \\
\midrule
\multicolumn{7}{l}{\textbf{Libraries 7--12 (2,019)}} \\
\quad Tafel term & 0.389 & 0.318 & 0.322 & 0.307 & 0.321 & 0.379 \\
\quad PR-ML (Ridge) & 0.223 & 0.173 & 0.172 & 0.183 & 0.211 & 0.322 \\
\quad Ridge & 0.166 & 0.209 & 0.168 & 0.205 & 0.230 & 0.239 \\
\quad PR-ML (ExtraTrees) & 0.242 & 0.182 & 0.180 & 0.174 & 0.206 & 0.281 \\
\quad ExtraTrees & 0.110 & 0.112 & 0.121 & 0.129 & 0.135 & 0.138 \\
\quad PR-ML (Ridge, $\alpha=10$) & 0.223 & 0.173 & 0.172 & 0.166 & 0.193 & 0.273 \\
\midrule
\multicolumn{7}{l}{\textbf{Larger of two}} \\
\quad Tafel term & 1.504 & 0.541 & 0.536 & 0.311 & 0.321 & 0.379 \\
\quad PR-ML (Ridge) & 1.519 & 0.194 & 0.329 & 0.230 & 0.211 & 0.322 \\
\quad Ridge & 1.693 & 1.330 & 0.943 & 1.098 & 1.118 & 1.131 \\
\quad PR-ML (ExtraTrees) & 1.478 & 0.416 & 0.457 & 0.259 & 0.206 & 0.281 \\
\quad ExtraTrees & 1.933 & 1.882 & 1.888 & 1.875 & 1.871 & 1.870 \\
\quad PR-ML (Ridge, $\alpha=10$) & 1.516 & 0.203 & 0.321 & 0.245 & 0.209 & 0.273 \\
\bottomrule
\end{tabular}
\end{table}

\begin{table}[!htbp]
\centering
\caption{\textbf{Reduced training sets in setting~1} (catalysts far above the training range). MAE
(\si{\milliamperepercmsquared}) of models fitted on 35 to 1,685 catalysts of libraries 1--5, with two (\textbf{a}) and three
(\textbf{b}) measured currents; median over 20 random subsets (1,685: the complete set, fitted
once). Hyperparameters were kept at the values selected on the complete set. A model was taken to have a
lower (higher) error when the 2.5th and 97.5th percentiles of the paired differences over the 20
random subsets were both below (above) zero. In every subset of
every reduced size, PR-ML (Ridge) and PR-ML (ExtraTrees) had a lower error than the same learner used
alone, and PR-ML (Ridge) had a lower error than the Tafel term. PR-ML (ExtraTrees) had a higher error
than the Tafel term with 35 and 50 training catalysts, no significant
difference with 85 and a lower error from 169 catalysts upward, so a linear residual learner is suited
to the smallest training sets.}
\label{tab:si_reduction}
\small
\setlength{\tabcolsep}{3pt}
\textbf{a}\par\smallskip
\begin{tabular}{rccccc}
\toprule
\makecell{Training\\catalysts} & \makecell{Tafel\\term} & \makecell{PR-ML\\(Ridge)} & Ridge & \makecell{PR-ML\\(ExtraTrees)} & ExtraTrees \\
\midrule
35 & 0.715 & 0.228 & 2.686 & 0.925 & 3.858 \\
50 & 0.728 & 0.207 & 2.723 & 0.850 & 3.798 \\
85 & 0.734 & 0.187 & 2.584 & 0.721 & 3.723 \\
169 & 0.748 & 0.192 & 2.454 & 0.631 & 3.592 \\
337 & 0.745 & 0.191 & 2.458 & 0.593 & 3.521 \\
674 & 0.730 & 0.184 & 2.465 & 0.572 & 3.432 \\
1,685 & 0.735 & 0.177 & 2.486 & 0.565 & 3.363 \\
\bottomrule
\end{tabular}

\bigskip
\textbf{b}\par\smallskip
\begin{tabular}{rccccc}
\toprule
\makecell{Training\\catalysts} & \makecell{Tafel\\term} & \makecell{PR-ML\\(Ridge)} & Ridge & \makecell{PR-ML\\(ExtraTrees)} & ExtraTrees \\
\midrule
35 & 0.599 & 0.282 & 2.122 & 0.782 & 3.836 \\
50 & 0.612 & 0.292 & 2.068 & 0.729 & 3.782 \\
85 & 0.618 & 0.285 & 1.924 & 0.631 & 3.697 \\
169 & 0.632 & 0.285 & 1.835 & 0.591 & 3.570 \\
337 & 0.629 & 0.289 & 1.695 & 0.551 & 3.487 \\
674 & 0.615 & 0.276 & 1.729 & 0.538 & 3.397 \\
1,685 & 0.620 & 0.269 & 1.721 & 0.532 & 3.338 \\
\bottomrule
\end{tabular}
\end{table}

\begin{table}[!htbp]
\centering
\caption{\textbf{Selection of catalysts above the training range by the current at
1.55~V$_\mathrm{RHE}$.} \textbf{a}, Each Ni--Pd--Pt--Ru library predicted from the other eleven
libraries. Selected: catalysts whose current at 1.55~V$_\mathrm{RHE}$ exceeds the highest
1.55~V$_\mathrm{RHE}$ current of the eleven training libraries. Above the training range: catalysts
whose current at 1.70~V$_\mathrm{RHE}$ exceeds the highest 1.70~V$_\mathrm{RHE}$ current of the
training libraries. \textbf{b}, MAE (\si{\milliamperepercmsquared}) at 1.70~V$_\mathrm{RHE}$ for the 206 selected catalysts of the
Ni--Ru library, with two, three and five currents measured per catalyst between 1.40 and
1.55~V$_\mathrm{RHE}$. PR-ML (Ridge) is the model of Table~\ref{tab:setting1} and
Fig.~\ref{fig:setting1}d; its residual learner uses the penalty weight of Ridge, the more accurate of
the two data-driven models on these catalysts. PR-ML (Ridge, $\alpha=10$) uses the penalty weight
selected for PR-ML itself. Lowest value in each column in bold.}
\label{tab:si_selection}
\small
\textbf{a}\par\smallskip
\setlength{\tabcolsep}{4.5pt}
\begin{tabular}{lcccccccccccc}
\toprule
Library & 1 & 2 & 3 & 4 & 5 & 6 (Ni--Ru) & 7 & 8 & 9 & 10 & 11 & 12 \\
\midrule
Catalysts & 339 & 331 & 339 & 336 & 340 & 322 & 339 & 338 & 340 & 326 & 336 & 340 \\
Selected & 0 & 0 & 0 & 0 & 0 & 206 & 0 & 0 & 0 & 0 & 0 & 0 \\
Above the training range & 0 & 0 & 0 & 0 & 0 & 186 & 0 & 0 & 0 & 0 & 0 & 0 \\
Selected and above & 0 & 0 & 0 & 0 & 0 & 185 & 0 & 0 & 0 & 0 & 0 & 0 \\
\bottomrule
\end{tabular}

\bigskip
\textbf{b}\par\smallskip
\setlength{\tabcolsep}{6pt}
\begin{tabular}{lcccccc}
\toprule
 & \multicolumn{3}{c}{\makecell{Fitted on\\libraries 1--5}} & \multicolumn{3}{c}{\makecell{Fitted on the other\\eleven libraries}} \\
\cmidrule(lr){2-4}\cmidrule(lr){5-7}
Number of measured currents & 2 & 3 & 5 & 2 & 3 & 5 \\
\midrule
Tafel term & 0.695 & 0.676 & 0.348 & 0.873 & 0.853 & 0.507 \\
Ridge & 1.938 & 1.320 & 1.593 & 1.466 & 1.336 & 1.487 \\
ExtraTrees & 2.685 & 2.668 & 2.645 & 2.314 & 2.290 & 2.274 \\
PR-ML (Ridge) & \textbf{0.188} & 0.390 & \textbf{0.271} & \textbf{0.429} & \textbf{0.508} & \textbf{0.278} \\
PR-ML (Ridge, $\alpha=10$) & 0.208 & \textbf{0.382} & 0.286 & 0.439 & 0.519 & 0.282 \\
\bottomrule
\end{tabular}
\end{table}

\begin{table}[!htbp]
\centering
\caption{\textbf{Setting 2 over the twelve Ni--Pd--Pt--Ru libraries.} MAE in $\log_{10}$ current at
1.70~V$_\mathrm{RHE}$, mean over the twelve libraries of the mean over 50 draws, with 5 and 20
labelled catalysts chosen at random or only from the less active half of each library
(range-shifted selection). Currents measured at 1.40 and 1.55~V$_\mathrm{RHE}$. GP, Gaussian process. PR-ML (GP, mean-residual prior mean) and PR-ML (GP, zero prior mean): PR-ML with a GP as
the residual learner, whose prior mean is the mean labelled residual (Fig.~\ref{fig:setting2}) or zero;
with a zero prior mean, the prediction returns to the Tafel term far from the labelled catalysts (as in
Table~\ref{tab:setting1}). Tafel term with a constant offset: the Tafel term plus the mean labelled
residual, which here equals the 1.55~V$_\mathrm{RHE}$ current multiplied by one fitted factor. Lowest
value in each column in bold. Bottom rows: number of libraries in which PR-ML (ExtraTrees) had the lower
error at five labelled catalysts, with the
Wilcoxon signed-rank test over libraries, and the mean difference in $\log_{10}$ current.}
\label{tab:si_twelve}
\small
\setlength{\tabcolsep}{4.5pt}
\begin{tabular}{lcccc}
\toprule
 & \multicolumn{2}{c}{Random selection} & \multicolumn{2}{c}{Range-shifted selection} \\
\cmidrule(lr){2-3}\cmidrule(lr){4-5}
Labelled catalysts & 5 & 20 & 5 & 20 \\
\midrule
PR-ML (ExtraTrees) & \textbf{0.036} & 0.026 & \textbf{0.041} & 0.034 \\
PR-ML (GP, mean-residual prior mean) & 0.038 & 0.027 & 0.044 & \textbf{0.033} \\
PR-ML (GP, zero prior mean) & 0.042 & \textbf{0.025} & 0.055 & 0.041 \\
Tafel term with a constant offset & 0.043 & 0.040 & 0.048 & 0.044 \\
ExtraTrees & 0.048 & 0.028 & 0.088 & 0.071 \\
Ridge & 0.052 & 0.027 & 0.078 & 0.042 \\
GP (composition and measured currents) & 0.053 & 0.026 & 0.092 & 0.071 \\
GP (composition) & 0.068 & 0.036 & 0.106 & 0.091 \\
\midrule
\multicolumn{5}{l}{PR-ML (ExtraTrees) against, at five labelled catalysts:} \\
\quad GP (composition) & \multicolumn{2}{c}{12 of 12, $p=4.9\times10^{-4}$; 0.032} & \multicolumn{2}{c}{12 of 12, $p=4.9\times10^{-4}$; 0.065} \\
\quad PR-ML (GP, mean-residual prior mean) & \multicolumn{2}{c}{11 of 12, $p=0.034$; 0.0017} & \multicolumn{2}{c}{10 of 12, $p=0.034$; 0.0032} \\
\bottomrule
\end{tabular}
\end{table}

\begin{table}[!htbp]
\centering
\caption{\textbf{Setting~2: PR-ML against the most accurate of three compared methods.} Measured
potentials 1.40 and 1.55~V$_\mathrm{RHE}$ or 1.50 and 1.55~V$_\mathrm{RHE}$; endpoint 1.70~V$_\mathrm{RHE}$.
Difference in MAE (in units of $10^{-3}$ in $\log_{10}$ current, mean over the twelve libraries) between
PR-ML and the most accurate compared method (negative: PR-ML is more accurate), with the 95\% confidence interval from 10,000 bootstrap resamples of the twelve libraries in
brackets. The most accurate compared method was chosen among the Tafel term (library slope), Ridge and
ExtraTrees by the mean error over the libraries, separately for each combination of measured potentials,
selection, number of labelled catalysts and group of catalysts; it was ExtraTrees (ET) or Ridge (R) in
every combination. \mbox{PR-ML} uses the same learner, features and labels as the compared method as its
residual learner, adds the full residual term and uses the Tafel term (library slope); it is therefore
PR-ML (ExtraTrees) when the compared method is ExtraTrees and PR-ML (Ridge) when it is Ridge. Below, above: catalysts below and above the largest label; all: all unlabelled catalysts.
$^\ast$, confidence interval below zero. In no combination is the compared method more accurate.}
\label{tab:si_stop155}
\small
\setlength{\tabcolsep}{4pt}
\begin{tabular}{llcccc}
\toprule
\makecell{Measured potentials\\(V$_\mathrm{RHE}$)} & Selection & \makecell{Labelled\\catalysts} & Below & Above & All \\
\midrule
1.40, 1.55 & random & 3 & \makecell[t]{$-12.92$ ET$^\ast$\\{}[$-21.10$, $-5.32$]} & \makecell[t]{$-38.46$ ET$^\ast$\\{}[$-47.83$, $-30.49$]} & \makecell[t]{$-24.54$ ET$^\ast$\\{}[$-33.62$, $-15.98$]} \\ \addlinespace[2pt]
 &  & 5 & \makecell[t]{$-5.58$ ET$^\ast$\\{}[$-10.30$, $-0.72$]} & \makecell[t]{$-5.97$ R$^\ast$\\{}[$-9.42$, $-2.88$]} & \makecell[t]{$-13.05$ ET$^\ast$\\{}[$-18.42$, $-7.91$]} \\ \addlinespace[2pt]
 &  & 10 & \makecell[t]{$-2.02$ ET\\{}[$-5.88$, $+1.86$]} & \makecell[t]{$-1.16$ R\\{}[$-2.54$, $+0.17$]} & \makecell[t]{$-1.32$ R$^\ast$\\{}[$-2.19$, $-0.55$]} \\ \addlinespace[2pt]
 &  & 20 & \makecell[t]{$-0.74$ ET\\{}[$-3.56$, $+2.08$]} & \makecell[t]{$-0.22$ R\\{}[$-0.98$, $+0.46$]} & \makecell[t]{$-0.35$ R$^\ast$\\{}[$-0.57$, $-0.16$]} \\ \addlinespace[2pt]
 &  & 40 & \makecell[t]{$-0.22$ ET\\{}[$-2.38$, $+1.93$]} & \makecell[t]{$-0.02$ R\\{}[$-0.48$, $+0.43$]} & \makecell[t]{$-0.56$ ET\\{}[$-2.71$, $+1.57$]} \\ \addlinespace[2pt]
 & range-shifted & 3 & \makecell[t]{$+2.06$ ET\\{}[$-5.95$, $+11.13$]} & \makecell[t]{$-82.45$ ET$^\ast$\\{}[$-111.31$, $-58.17$]} & \makecell[t]{$-52.68$ ET$^\ast$\\{}[$-72.87$, $-35.47$]} \\ \addlinespace[2pt]
 &  & 5 & \makecell[t]{$+2.29$ ET\\{}[$-4.01$, $+9.01$]} & \makecell[t]{$-12.20$ R$^\ast$\\{}[$-15.62$, $-9.01$]} & \makecell[t]{$-8.72$ R$^\ast$\\{}[$-11.08$, $-6.30$]} \\ \addlinespace[2pt]
 &  & 10 & \makecell[t]{$-0.84$ R$^\ast$\\{}[$-1.45$, $-0.31$]} & \makecell[t]{$-3.29$ R$^\ast$\\{}[$-4.51$, $-2.09$]} & \makecell[t]{$-2.22$ R$^\ast$\\{}[$-3.03$, $-1.38$]} \\ \addlinespace[2pt]
 &  & 20 & \makecell[t]{$-0.14$ R$^\ast$\\{}[$-0.27$, $-0.03$]} & \makecell[t]{$-1.44$ R$^\ast$\\{}[$-2.42$, $-0.65$]} & \makecell[t]{$-0.86$ R$^\ast$\\{}[$-1.41$, $-0.42$]} \\ \addlinespace[2pt]
 &  & 40 & \makecell[t]{$+0.50$ ET\\{}[$-1.38$, $+2.49$]} & \makecell[t]{$-0.54$ R$^\ast$\\{}[$-1.11$, $-0.09$]} & \makecell[t]{$-0.31$ R$^\ast$\\{}[$-0.64$, $-0.06$]} \\ \addlinespace[2pt]
\midrule
1.50, 1.55 & random & 3 & \makecell[t]{$-12.68$ ET$^\ast$\\{}[$-21.40$, $-3.71$]} & \makecell[t]{$-36.20$ ET$^\ast$\\{}[$-43.66$, $-28.55$]} & \makecell[t]{$-23.29$ ET$^\ast$\\{}[$-33.17$, $-13.72$]} \\ \addlinespace[2pt]
 &  & 5 & \makecell[t]{$-4.58$ ET$^\ast$\\{}[$-8.71$, $-0.14$]} & \makecell[t]{$-3.81$ R$^\ast$\\{}[$-6.35$, $-1.64$]} & \makecell[t]{$-10.37$ ET$^\ast$\\{}[$-14.96$, $-5.63$]} \\ \addlinespace[2pt]
 &  & 10 & \makecell[t]{$-1.44$ R$^\ast$\\{}[$-2.72$, $-0.06$]} & \makecell[t]{$-2.19$ R$^\ast$\\{}[$-3.28$, $-1.21$]} & \makecell[t]{$-1.58$ R$^\ast$\\{}[$-2.83$, $-0.33$]} \\ \addlinespace[2pt]
 &  & 20 & \makecell[t]{$-0.60$ R\\{}[$-1.18$, $+0.04$]} & \makecell[t]{$-1.08$ R$^\ast$\\{}[$-1.84$, $-0.39$]} & \makecell[t]{$-0.63$ R$^\ast$\\{}[$-1.19$, $-0.01$]} \\ \addlinespace[2pt]
 &  & 40 & \makecell[t]{$-0.18$ R\\{}[$-0.43$, $+0.12$]} & \makecell[t]{$-0.30$ R\\{}[$-0.97$, $+0.35$]} & \makecell[t]{$-0.18$ R\\{}[$-0.43$, $+0.10$]} \\ \addlinespace[2pt]
 & range-shifted & 3 & \makecell[t]{$+1.80$ ET\\{}[$-5.93$, $+10.50$]} & \makecell[t]{$-81.65$ ET$^\ast$\\{}[$-109.41$, $-59.65$]} & \makecell[t]{$-51.78$ ET$^\ast$\\{}[$-69.96$, $-36.40$]} \\ \addlinespace[2pt]
 &  & 5 & \makecell[t]{$+1.97$ ET\\{}[$-3.64$, $+8.61$]} & \makecell[t]{$-7.87$ R$^\ast$\\{}[$-13.15$, $-1.89$]} & \makecell[t]{$-5.78$ R$^\ast$\\{}[$-9.31$, $-2.25$]} \\ \addlinespace[2pt]
 &  & 10 & \makecell[t]{$-1.11$ R$^\ast$\\{}[$-1.72$, $-0.52$]} & \makecell[t]{$-2.21$ R\\{}[$-5.35$, $+1.89$]} & \makecell[t]{$-1.72$ R\\{}[$-3.57$, $+0.50$]} \\ \addlinespace[2pt]
 &  & 20 & \makecell[t]{$-0.42$ R$^\ast$\\{}[$-0.69$, $-0.16$]} & \makecell[t]{$-0.23$ R\\{}[$-2.42$, $+2.92$]} & \makecell[t]{$-0.31$ R\\{}[$-1.51$, $+1.39$]} \\ \addlinespace[2pt]
 &  & 40 & \makecell[t]{$-0.11$ R$^\ast$\\{}[$-0.20$, $-0.03$]} & \makecell[t]{$-0.10$ R\\{}[$-1.48$, $+1.64$]} & \makecell[t]{$-0.10$ R\\{}[$-0.94$, $+0.91$]} \\ \addlinespace[2pt]
\bottomrule
\end{tabular}
\end{table}

\begin{table}[!htbp]
\centering
\caption{\textbf{All combinations of measurement design and endpoint in the FastCat and DASH tests
above the label range.} Number of paired comparisons in which PR-ML with each residual learner had a
lower error / no significant difference / a higher error than the same learner used alone or than the
Tafel term (dataset median slope). Differences were judged with 95\% confidence intervals from 2,000
bootstrap resamples of the evaluated catalysts.
Six measurement designs (overpotentials at 1 and 2; 1 and 5; 2 and 5; 1, 2 and 5; 5 and 10; or 1, 2, 5
and 10~\si{\milliamperepercmsquared}), each combined with every endpoint ($\eta_{10}$, $\eta_{20}$,
$\eta_{50}$) above its highest measured current density, give 16 combinations. With two label
fractions and five numbers of labelled catalysts, there are 160 comparisons for each learner and
dataset. $^{\dagger}$The GP was run for the two designs of the main text (20 comparisons). Ridge uses composition and
measured values. In DASH, the Tafel term, which uses no endpoint labels, had a lower error than PR-ML in
most comparisons.}
\label{tab:si_combinations}
\small
\setlength{\tabcolsep}{3.5pt}
\begin{tabular}{lcccc}
\toprule
 & \multicolumn{2}{c}{vs the same learner used alone} & \multicolumn{2}{c}{vs the Tafel term} \\
\cmidrule(lr){2-3}\cmidrule(lr){4-5}
Residual learner & FastCat & DASH & FastCat & DASH \\
\midrule
ExtraTrees & 132 / 19 / 9 & 90 / 70 / 0 & 39 / 66 / 55 & 10 / 15 / 135 \\
GP$^{\dagger}$ & 19 / 1 / 0 & 15 / 5 / 0 & 7 / 4 / 9 & 0 / 0 / 20 \\
Ridge & 103 / 40 / 17 & 104 / 44 / 12 & 28 / 43 / 89 & 0 / 0 / 160 \\
Ridge (measured values only) & 133 / 27 / 0 & 62 / 37 / 61 & 31 / 77 / 52 & 0 / 17 / 143 \\
\bottomrule
\end{tabular}
\end{table}

\begin{table}[!htbp]
\centering
\caption{\textbf{FastCat and DASH with the catalysts divided by the order in which they were made.}
FastCat: the labelled catalysts were drawn from the 60 earliest of the 89 catalysts, and the 29 later
catalysts were evaluated. DASH: the labelled catalysts were drawn from synthesis batches 1 and 2 (51
catalysts), and the 84 catalysts of batches 3--5 were evaluated. \textbf{a}, Number of paired
comparisons in which PR-ML with each residual learner had a lower error / no significant difference
/ a higher error than the same learner used alone, for the two designs of the main text and 3, 5, 10, 20
or 40 labelled catalysts (200 draws each). Columns give the results for all later catalysts and for the
later catalysts more active than the best labelled catalyst. \textbf{b}, The same over all 16
combinations of design and endpoint, without the GP. \textbf{c}, MAE (mV) of all later catalysts with
five labelled catalysts, for PR-ML with the residual learner / the learner used alone; last row, the
Tafel term (dataset median slope), which uses no endpoint labels. For all later
catalysts in DASH, the Tafel term was more accurate than PR-ML in 28 of 40 comparisons (four residual learners, the two designs of the
main text and five numbers of labelled catalysts) and showed no significant difference in the other 12; in FastCat, PR-ML was more accurate than the Tafel term in
20 of 40 and less accurate in 4.}
\label{tab:si_chrono}
\small
\textbf{a}\par\smallskip
\setlength{\tabcolsep}{4pt}
\begin{tabular}{lcccc}
\toprule
 & \multicolumn{2}{c}{FastCat} & \multicolumn{2}{c}{DASH} \\
\cmidrule(lr){2-3}\cmidrule(lr){4-5}
Residual learner & \makecell{all later\\catalysts} & \makecell{more active than the\\best labelled catalyst} & \makecell{all later\\catalysts} & \makecell{more active than the\\best labelled catalyst} \\
\midrule
ExtraTrees & 3 / 7 / 0 & 7 / 3 / 0 & 5 / 5 / 0 & 7 / 3 / 0 \\
GP & 2 / 8 / 0 & 5 / 5 / 0 & 7 / 3 / 0 & 8 / 2 / 0 \\
Ridge & 8 / 2 / 0 & 7 / 2 / 1 & 8 / 2 / 0 & 8 / 2 / 0 \\
Ridge (measured values only) & 6 / 4 / 0 & 5 / 5 / 0 & 4 / 5 / 1 & 3 / 6 / 1 \\
\bottomrule
\end{tabular}

\bigskip
\textbf{b}\par\smallskip
\begin{tabular}{lcccc}
\toprule
 & \multicolumn{2}{c}{FastCat} & \multicolumn{2}{c}{DASH} \\
\cmidrule(lr){2-3}\cmidrule(lr){4-5}
Residual learner & \makecell{all later\\catalysts} & \makecell{more active than the\\best labelled catalyst} & \makecell{all later\\catalysts} & \makecell{more active than the\\best labelled catalyst} \\
\midrule
ExtraTrees & 12 / 54 / 14 & 43 / 29 / 8 & 27 / 48 / 5 & 47 / 31 / 2 \\
Ridge & 30 / 42 / 8 & 30 / 40 / 10 & 62 / 12 / 6 & 58 / 20 / 2 \\
Ridge (measured values only) & 45 / 28 / 7 & 41 / 31 / 8 & 37 / 18 / 25 & 38 / 20 / 22 \\
\bottomrule
\end{tabular}

\bigskip
\textbf{c}\par\smallskip
\begin{tabular}{lcccc}
\toprule
 & \multicolumn{2}{c}{FastCat} & \multicolumn{2}{c}{DASH} \\
\cmidrule(lr){2-3}\cmidrule(lr){4-5}
\makecell[l]{Residual learner\\(PR-ML / learner used alone)} & $\eta_{10}$ & $\eta_{50}$ & $\eta_{10}$ & $\eta_{50}$ \\
\midrule
ExtraTrees & 43.4 / 54.5 & 32.2 / 55.1 & 8.7 / 19.1 & 18.0 / 33.4 \\
GP & 45.4 / 50.9 & 32.9 / 57.0 & 9.3 / 20.6 & 21.5 / 43.9 \\
Ridge & 53.4 / 73.2 & 45.1 / 70.4 & 11.6 / 30.8 & 24.2 / 36.9 \\
Ridge (measured values only) & 36.6 / 52.9 & 33.2 / 40.1 & 15.7 / 25.1 & 22.3 / 24.0 \\
\midrule
Tafel term (dataset median slope) & 58.1 & 32.1 & 8.4 & 16.3 \\
\bottomrule
\end{tabular}
\end{table}

\clearpage
\phantomsection\label{note:si_dash}
\section*{\raggedright Supplementary Note 1: Prediction of the current at a higher potential in DASH}

\paragraph{Purpose.} The Ni--Pd--Pt--Ru results show prediction of the current at a higher
potential for catalysts above the training range. This note tests the same situation in a second
dataset, DASH, in which the catalysts were prepared in five synthesis batches (34, 37, 52, 64 and 64
of the 251 curves) and each catalyst has a linear sweep voltammogram in 0.5~M H$_2$SO$_4$.

\paragraph{Data.} The currents of 251 DASH curves were read at fixed potentials. The recorded current
stops increasing near 10~mA (51.0--51.9~\si{\milliamperepercmsquared}); 169 curves reach this value,
and currents at and above the potential where it is first reached were treated as lower bounds.
Catalysts whose endpoint lay at or above this current were not used as labels and were evaluated only
by how far the prediction fell below the lower bound.

\paragraph{Design.} Eight combinations of measured potentials and endpoint were tested: currents
measured at 1.30 and 1.35,
1.35 and 1.40, or 1.30 and 1.40~V$_\mathrm{RHE}$ with endpoints at 1.45 and 1.50~V$_\mathrm{RHE}$,
and currents measured at 1.40 and 1.45 or 1.35 and 1.45~V$_\mathrm{RHE}$ with the endpoint at
1.50~V$_\mathrm{RHE}$. Within each batch, catalysts were sorted by the potential at which their curve
first reached 5~\si{\milliamperepercmsquared}. With the 50/50 split, the labelled catalysts were taken from the less active half and the more active
half was predicted; with the 30/30 split, they were taken from the least active 30\% and the most active
30\% was predicted. The models were fitted separately in each batch, and the predicted catalysts of the
five batches were pooled for evaluation.

\paragraph{Choices made on the labelled catalysts only.} In each batch, three models were chosen using
only the labelled catalysts of the batch: (i) Tafel extrapolation with a selected slope, that is, the extrapolation of
Eq.~\ref{eq:tafel} with the slope that gave the lowest error on the labelled catalysts among eight options (the median two-point slope of the batch, the
catalyst's own slope, standard Tafel slopes of 40, 60, 80 and 120~mV~dec$^{-1}$, the median slope of the less
active catalysts, and the combination used in setting~1, 0.75 times this median plus 0.25 times the
catalyst's own slope);
(ii) the data-driven model (selected by cross-validation), that is, the learner with the lowest
five-fold cross-validation (CV) error among five learners (Ridge and ExtraTrees on composition and
measured currents, Ridge on the measured currents only, and Ridge and ExtraTrees on composition only);
and (iii) PR-ML (CV-selected learner) with a selected slope, that is, PR-ML with the learner of (ii) as
the residual learner, the slope chosen among the same eight options by its own
five-fold cross-validation, and the residual term halved and reduced for catalysts far from the labelled
catalysts as in setting~1 ($\beta=0.5$, $\lambda=1$, $q=0.95$). The design and the comparisons were fixed
before the results were seen. Differences were judged with 95\% confidence
intervals from 2,000 bootstrap resamples of the predicted catalysts.

\paragraph{Results.} Supplementary Table~\ref{tab:si_dash} summarizes the eight combinations.
\begin{itemize}
\item With the 50/50 split, for the catalysts above the label range, PR-ML (CV-selected learner) with a
selected slope had a lower error than the Tafel extrapolation with a selected slope in 6 of 8 combinations and no significant difference in 2. Compared
with the data-driven model, it had a lower error in 1 combination, no significant difference in 6 and a
higher error in 1 (currents measured at 1.40 and 1.45~V$_\mathrm{RHE}$, endpoint 1.50~V$_\mathrm{RHE}$;
difference 0.023 in $\log_{10}$ current).
\item With the 30/30 split, this PR-ML model had a lower error than the data-driven model in 7 of 8 combinations
and no significant difference in 1. Compared with the Tafel extrapolation, it had a lower error in 3
combinations, no significant difference in 2 and a higher error in 3.
\item In the two combinations with currents measured up to 1.45~V$_\mathrm{RHE}$ and the endpoint at
1.50~V$_\mathrm{RHE}$, with the 30/30 split, this PR-ML model had a lower error than both compared methods
for the catalysts above the label range: MAE in $\log_{10}$ current 0.188 and 0.175, against 0.269 and
0.283 for the Tafel extrapolation and 0.462 and 0.722 for the data-driven model.
\end{itemize}
With both splits, this PR-ML model had the lowest median MAE of the three
methods for the
catalysts above the label range (0.223 and 0.293 in $\log_{10}$ current), and in no combination was it
less accurate than both compared methods.

\paragraph{Slopes and learners chosen.} The slope chosen for this PR-ML model was a standard Tafel slope (40, 60, 80 or 120~mV~dec$^{-1}$) in 35 of the 40 fits (five batches and eight combinations) with the
50/50 split and in 37 of 40 with the 30/30 split, most often 80~mV~dec$^{-1}$ (21 and 20 fits). The learner chosen most often for the data-driven model was Ridge on the measured currents only (26
and 27 of 40 fits), followed by ExtraTrees on composition and measured currents (13 and 9). Because the
slope was chosen from these eight options, this test does not use the slope of the Tafel term of the main
text, which is estimated from the measured curves.

\begin{table}[!htbp]
\centering
\caption{\textbf{Prediction of the current at a higher potential in DASH: PR-ML (\mbox{CV-selected} learner) with a selected slope against the two compared methods.} Number of the eight combinations in which this
model had a lower error / no significant difference / a higher error than each compared method, and the
median difference in MAE ($\log_{10}$ current; negative, this model is more accurate). More accurate compared
method: whichever of the Tafel extrapolation with a selected slope and the data-driven model had the lower error on the pooled
predicted catalysts of each combination and group of catalysts. Above the label range: predicted catalysts whose current exceeds
the largest label; above by more than one IQR: catalysts whose logarithmic current exceeds the largest
logarithmic label by more than one interquartile range of the logarithmic labels. Bottom rows: median
MAE over the eight combinations for the catalysts above the label range. Catalysts predicted per combination (five
batches pooled): with the 50/50 split, 114 with a measured endpoint at 1.45~V$_\mathrm{RHE}$, of which
100 lie above the label range and 13 above it by more than one IQR, and 46--49 at 1.50~V$_\mathrm{RHE}$,
of which 28--31 lie above the label range and 6 above it by more than one IQR; with the 30/30 split, 63
at 1.45~V$_\mathrm{RHE}$, of which 62 lie above the label range and 37 above it by more than one IQR,
and 18--19 at 1.50~V$_\mathrm{RHE}$, of which 16--17 lie above the label range and 12--13 above it by
more than one IQR. For the catalysts above the label range, the more accurate compared method was the
data-driven model in 7 of 8 combinations with the 50/50 split and the Tafel extrapolation in 7 of 8 with the
30/30 split.}
\label{tab:si_dash}
\small
\setlength{\tabcolsep}{4pt}
\begin{tabular}{llccc}
\toprule
Split & Predicted catalysts & \makecell{vs Tafel\\extrapolation} & vs data-driven model & \makecell{vs more accurate\\compared method} \\
\midrule
50/50 & above the label range & 6 / 2 / 0 ($-0.049$) & 1 / 6 / 1 ($-0.024$) & 0 / 7 / 1 ($-0.013$) \\
50/50 & above by more than one IQR & 5 / 1 / 2 ($-0.026$) & 2 / 6 / 0 ($-0.005$) & 2 / 4 / 2 ($+0.003$) \\
30/30 & above the label range & 3 / 2 / 3 ($+0.016$) & 7 / 1 / 0 ($-0.251$) & 2 / 3 / 3 ($+0.020$) \\
30/30 & above by more than one IQR & 2 / 2 / 4 ($-0.004$) & 7 / 1 / 0 ($-0.268$) & 1 / 3 / 4 ($+0.033$) \\
\midrule
\multicolumn{2}{l}{Median MAE, above the label range} & Tafel extrapolation & Data-driven model & \makecell{PR-ML (CV-selected learner)\\with a selected slope} \\
50/50 & & 0.275 & 0.250 & 0.223 \\
30/30 & & 0.302 & 0.561 & 0.293 \\
\bottomrule
\end{tabular}

\end{table}

\end{document}